%% file: main.tex
\documentclass[10pt,twocolumn,letterpaper]{article}

\usepackage[pagenumbers]{cvpr} 

\input{preamble}

\definecolor{cvprblue}{rgb}{0.21,0.49,0.74}
\usepackage[pagebackref,breaklinks,colorlinks,allcolors=cvprblue]{hyperref}

\def\paperID{194} 
\def\confName{CVPR}
\def\confYear{2026}

\title{Unleashing Temporal Capacity of Spiking Neural Networks through Spatiotemporal Separation}

\author{Yiting Dong$^{1,2}$, Zhaofei Yu$^{2,3}$\thanks{Corresponding author} , Jianhao Ding$^{1,2}$, Zijie Xu$^{2,3}$, Tiejun Huang$^{1,2,3}$\\
$^1$School of Computer Science, Peking University\\
$^2$State Key Laboratory of Multimedia Information Processing, Peking University\\
$^3$Institute for Artificial Intelligence, Peking University\\
{\tt\small \{dongyiting, yuzf12, 2506398078, tjhuang\}@pku.edu.cn \{zjxu25\}@stu.pku.edu.cn} 
}

\begin{document}
\maketitle
\input{sec/0_abstract}    
\input{sec/1_introduction}
\input{sec/2_related_work}
\input{sec/3_methods}

\input{sec/4_experiments}

\input{sec/5_conclusion}
{
    \small
    \bibliographystyle{ieeenat_fullname}
    \bibliography{main}
}

\input{sec/X_suppl}

\end{document}

%% file: preamble.tex
\usepackage{microtype}

\renewcommand{\paragraph}[1]{\vspace{.3em}\noindent\textbf{#1.}}

\usepackage{booktabs}
\usepackage{subcaption}
\usepackage{multirow}
\usepackage{array}
\usepackage[table,xcdraw]{xcolor}
\usepackage{colortbl}
\usepackage{arydshln}
\usepackage[normalem]{ulem}
\useunder{\uline}{\ul}{}


%% file: sec/0_abstract.tex
\begin{abstract}
Spiking Neural Networks (SNNs) are considered naturally suited for temporal processing, with membrane potential propagation widely regarded as the core temporal modeling mechanism. However, existing research lack analysis of its actual contributions in complex temporal tasks. We design Non-Stateful (NS) models progressively removing membrane propagation to quantify its stage-wise role. Experiments reveal a counterintuitive phenomenon: moderate removal in shallow or deep layers improves performance, while excessive removal causes collapse. We attribute this to spatio-temporal resource competition where neurons encode both semantics and dynamics within limited range, with temporal state consuming capacity for spatial learning. Based on this, we propose Spatial-Temporal Separable Network (STSep), decoupling residual blocks into independent spatial and temporal branches. The spatial branch focuses on semantic extraction while the temporal branch captures motion through explicit temporal differences. Experiments on Something-Something V2, UCF101, and HMDB51 show STSep achieves superior performance, with retrieval task and attention analysis confirming focus on motion rather than static appearance. This work provides new perspectives on SNNs' temporal mechanisms and an effective solution for spatiotemporal modeling in video understanding. 
\vspace{-1.em}
\end{abstract}

%% file: sec/1_introduction.tex
\section{Introduction}
\vspace{-0.5em}

Spiking Neural Networks (SNNs) have garnered significant attention due to their spike-driven temporal dynamics and biologically plausible modeling of neural systems\cite{maass1997networks,gerstner2002spiking}, demonstrating potential in neuromorphic computing\cite{orchard2015converting,roy2019towards}, speech recognition\cite{wu2020deep,wu2018biologically}, and low-power edge devices \cite{davies2018loihi,akopyan2015truenorth}. Spiking neurons accumulate input currents to update membrane potential, emitting spikes and resetting when the potential exceeds a threshold\cite{burkitt2006review,gerstner2002spiking}. This dynamic process naturally enables membrane potential to serve as a carrier for information propagation across time steps\cite{gerstner2014neuronal,gerstner2002spiking}. From a computational graph perspective, SNNs can be viewed as recurrent architectures where temporal states evolve recursively over time, forming a state-space model $V_t = f(V_{t-1}, I_t)$. This mechanism enables SNNs to achieve temporal modeling capabilities without introducing additional parameters such as 3D convolutions\cite{tran2015learning,carreira2017quo}, theoretically providing intrinsic advantages for processing temporal data\cite{fang2021incorporating}. However, despite membrane potential propagation being widely considered fundamental to temporal modeling in SNNs\cite{wu2018spatio,zheng2021going,fang2021incorporating}, existing research rarely investigates the actual contributions of temporal state across different network layers or provides systematic analysis of its effectiveness in complex temporal tasks.

\begin{figure}[t]
  \centering
  \vspace{-0.5em}
   \includegraphics[width=0.75\linewidth]{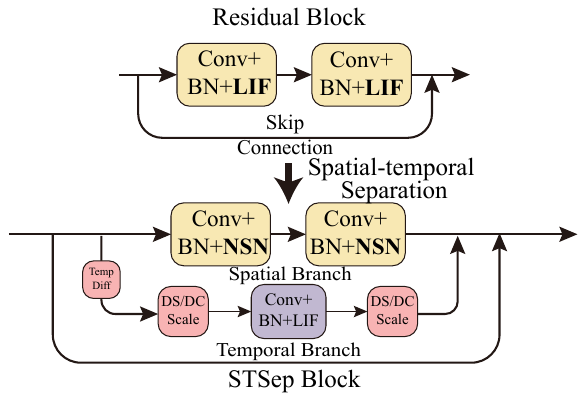}
   \vspace{-0.5em}
   \caption{\textbf{The spatiotemporal separable network.} Separating residual blocks into non-stateful spatial and temporal branches.}
   \label{fig:separation}
   \vspace{-1.8em}
\end{figure}

On one hand, existing research predominantly evaluates SNNs' temporal modeling capabilities on event camera datasets such as DVS128 Gesture\cite{amir2017low}, DVS-CIFAR10\cite{li2017cifar10}, and N-Caltech101\cite{orchard2015converting}. However, these datasets are limited in scale and temporal complexity, insufficient for validating models' ability to process intricate temporal patterns\cite{pfeiffer2018deep}. In contrast, the video understanding domain has established large-scale annotated datasets such as Something-Something\cite{goyal2017something} and Kinetics\cite{carreira2017quo}, which contains rich temporal dependencies and dynamic interactions, widely used to assess temporal modeling capabilities\cite{wang2016temporal,feichtenhofer2019slowfast}. We argue that evaluation on such large-scale video datasets can genuinely reveal SNNs' capacity for complex temporal processing.  On the other hand, temporal data inherently couples spatial information (appearance, shape contours) with temporal information (motion, positional changes)\cite{simonyan2014two}. However, deep networks commonly exhibit spatial bias, tending to rely on static spatial features rather than temporal dynamics\cite{choi2019can,goyal2017something}. SNN architectures, built upon image recognition structures like ResNet\cite{he2016deep} and VGG\cite{simonyan2014very}, inherit this spatial preference\cite{fang2021deep,sengupta2019going}, struggling to effectively extract temporal information from spatiotemporally mixed video data and thus suppressing temporal modeling capacity.

To investigate this issue, we design a series of Non-Stateful (NS) models analyzing membrane potential propagation across network layers. NS progressively remove temporal state stage-wise, transforming SNNs into non-stateful variants to quantify temporal state's contribution at each stage. Surprisingly, results reveal a counterintuitive phenomenon that performance does not decline monotonically with progressive removal (See Section \ref{sec_exp:NS}). Eliminating modest temporal state in bottom or top layers improves performance, while further removal causes sharp degradation. We attribute this non-monotonic pattern to \textbf{spatio-temporal resource competition}. In SNNs, neurons need encode both spatial semantics and temporal dynamics within limited dynamic range\cite{fang2021incorporating}. While state maintenance enables temporal integration, it consumes representational capacity for spatial learning. Moderate temporal state release liberates resources, enabling more effective spatial feature extraction. Ablation experiments show this phenomenon is pronounced in bottom and top layers, which handle initial encoding and final semantic extraction\cite{zeiler2014visualizing}, most sensitive to capacity demands. These findings yield two insights: \texttt{(i)} spatial and temporal information should be decoupled to avoid resource competition. \texttt{(ii)} not all layers require temporal state. Different depths should adopt different temporal modeling strategies.  
 
Based on these insights, we propose \textbf{Spatial-Temporal Separable Network (STSep)}, explicitly decoupling residual blocks into independent spatial and temporal branches processing spatial semantics and temporal dynamics respectively (See Figure \ref{fig:separation}). The spatial branch omits membrane potential propagation, dedicating full capacity to spatial feature extraction. The temporal branch explicitly computes temporal differences to capture motion patterns, providing direct temporal cues. Through progressive separation experiments on Something-Something V2\cite{goyal2017something}, we identify optimal separation configurations and validate the mechanism's effectiveness via ablations studies. We further reproduce multiple SNN temporal modeling methods\cite{fang2021incorporating,deng2022temporal,zhang2019spike,dong2024temporal,zhu2024tcja,zheng2021going}. Experiments on Something-Something V2\cite{goyal2017something}, UCF101\cite{soomro2012ucf101}, and HMDB51\cite{kuehne2011hmdb} show STSep achieves superior performance across all methods. Attention heatmap analysis reveals STSep effectively attends to motion regions rather than static appearance. Finally, video retrieval tasks further validate STSep's superior temporal semantic extraction.

%% file: sec/2_related_work.tex
\section{Related Work}
\vspace{-0.8em}

\paragraph{Video Understanding with 2D and 3D Networks}
Video understanding requires temporal modeling beyond spatial feature extraction. Two-stream networks \cite{simonyan2014two,feichtenhofer2016convolutional} model appearance and motion via RGB and optical-flow streams, establishing explicit motion representation. TSN \cite{wang2016temporal,wang2018temporal} achieves long-range temporal modeling through sparse sampling and segment consensus. Building on 2D convolution(e.g. TSM, TEA, TEINet) \cite{lin2019tsm,li2020tea,liu2020teinet,zhou2018temporal}, methods add temporal modules to enhance motion dependencies features through optical flow, displacement, or attention \cite{liu2020teinet,li2020tea}.
 
3D convolutional networks learn motion patterns via joint spatio-temporal convolutions \cite{tran2015learning,carreira2017quo,feichtenhofer2019slowfast,tran2018closer,qiu2017learning,xie2018rethinking,feichtenhofer2020x3d,zolfaghari2018eco}. Building on C3D \cite{tran2015learning}, subsequent methods explore reusing large-scale image pretraining (I3D \cite{carreira2017quo}), improving training efficiency by factorizing 3D convolutions into spatial-temporal cascades (R(2+1)D \cite{tran2018closer}, S3D \cite{xie2018rethinking}, P3D \cite{qiu2017learning}), combining spatial semantics and fast motion (SlowFast \cite{feichtenhofer2019slowfast}), and designing temporal structures (X3D \cite{feichtenhofer2020x3d}, ECO \cite{zolfaghari2018eco}). However, the joint spatio-temporal modeling of 3D convolutions incurs dense floating-point operations, leading to high computational cost and energy consumption.

\begin{table}[] 
\centering

\setlength{\tabcolsep}{2.5pt}
\renewcommand{\arraystretch}{1.1}
\scalebox{0.89}{


\begin{tabular}{l|cccc}
\multicolumn{1}{c|}{\textbf{methods}}  & \textbf{\begin{tabular}[c]{@{}c@{}}Membrane\\ Potential\end{tabular}} & \textbf{\begin{tabular}[c]{@{}c@{}}Recurrent\\ Connection\end{tabular}} & \textbf{\begin{tabular}[c]{@{}c@{}}Training\\ Strategy\end{tabular}} & \textbf{\begin{tabular}[c]{@{}c@{}}Feature\\ Interaction\end{tabular}} \\ \hline
PLIF \cite{fang2021incorporating}      & $\checkmark$                                                          &                                                                         &                                                                      &                                                                        \\
\hdashline RSNN \cite{zhang2019spike}  &                                                                       & $\checkmark$                                                            &                                                                      &                                                                        \\
\hdashline TET \cite{deng2022temporal} &                                                                       &                                                                         & $\checkmark$                                                         &                                                                        \\
TKS \cite{dong2024temporal}            &                                                                       &                                                                         & $\checkmark$                                                         &                                                                        \\
\hdashline TDBN \cite{zheng2021going}  &                                                                       &                                                                         &                                                                      & $\checkmark$                                                           \\
TCJA \cite{zhu2024tcja}                &                                                                       &                                                                         &                                                                      & $\checkmark$                                                          
\end{tabular}

}
\vspace{-0.5em}
\caption{\textbf{Summary of related works on temporal dynamics in SNNs from four perspectives:} Membrane Potential, Recurrent Connection, Training Strategy, and Feature Interaction.}
\vspace{-1.8em}
\label{tab:temporal_methods}
\end{table}

\paragraph{Temporal Dynamics in Spiking Neural Networks}
Spiking neural networks (SNNs) naturally process temporal information through firing spike (See Table \ref{tab:temporal_methods}). \textit{(i) Membrane potential perspective:} LIF neurons realize temporal dynamics via leakage and threshold-based firing \cite{gerstner2002spiking}, and parameterized variants such as PLIF \cite{fang2021incorporating} enhance temporal representation through learnable time constants. \textit{(ii) Recurrent perspective:} RSNN \cite{zhang2019spike} introduces recurrent connections to enhance temporal dependencies, while other works incorporate lateral or top-down feedback connections \cite{cheng2020lisnn}. \textit{(iii) Training perspective:} TET \cite{deng2022temporal} employs temporal consistency regularization to enhance temporal feature coherence, while TKS \cite{dong2024temporal} utilizes temporal knowledge self-distillation. \textit{(iV)  Feature perspective:} TDBN \cite{zheng2021going} introduces temporal difference blocks to capture motion cues, and Tcja \cite{zhu2024tcja} employs temporal joint channel attention to enhance temporal feature interactions.


\paragraph{Temporal Difference for Motion Modeling}
Temporal differencing is a fundamental operation for capturing motion \cite{zhang1993automatic,cutler2002robust,huwer2000adaptive}, traditionally implemented by computing pixel differences between adjacent frames in early video analysis.
It also underlies modern deep architectures: ResNet residuals can be viewed as first-order feature differencing \cite{he2016deep}, while TSM \cite{lin2019tsm} achieves implicit comparison via temporal feature shifts.
Optical flow networks such as FlowNet and PWC-Net \cite{dosovitskiy2015flownet,sun2018pwc} learn dense frame-to-frame correspondences.
More direct approaches like TDN \cite{wang2021tdn} apply multi-scale temporal differencing to capture motion at multiple scales, and TRN \cite{zhou2018temporal} or MotionSqueeze \cite{kwon2020motionsqueeze} enhance temporal learning via feature-relation modeling.

%% file: sec/3_methods.tex
\section{Non-stateful Networks}

\begin{figure}[t]
    \centering
    \includegraphics[width=0.9\linewidth]{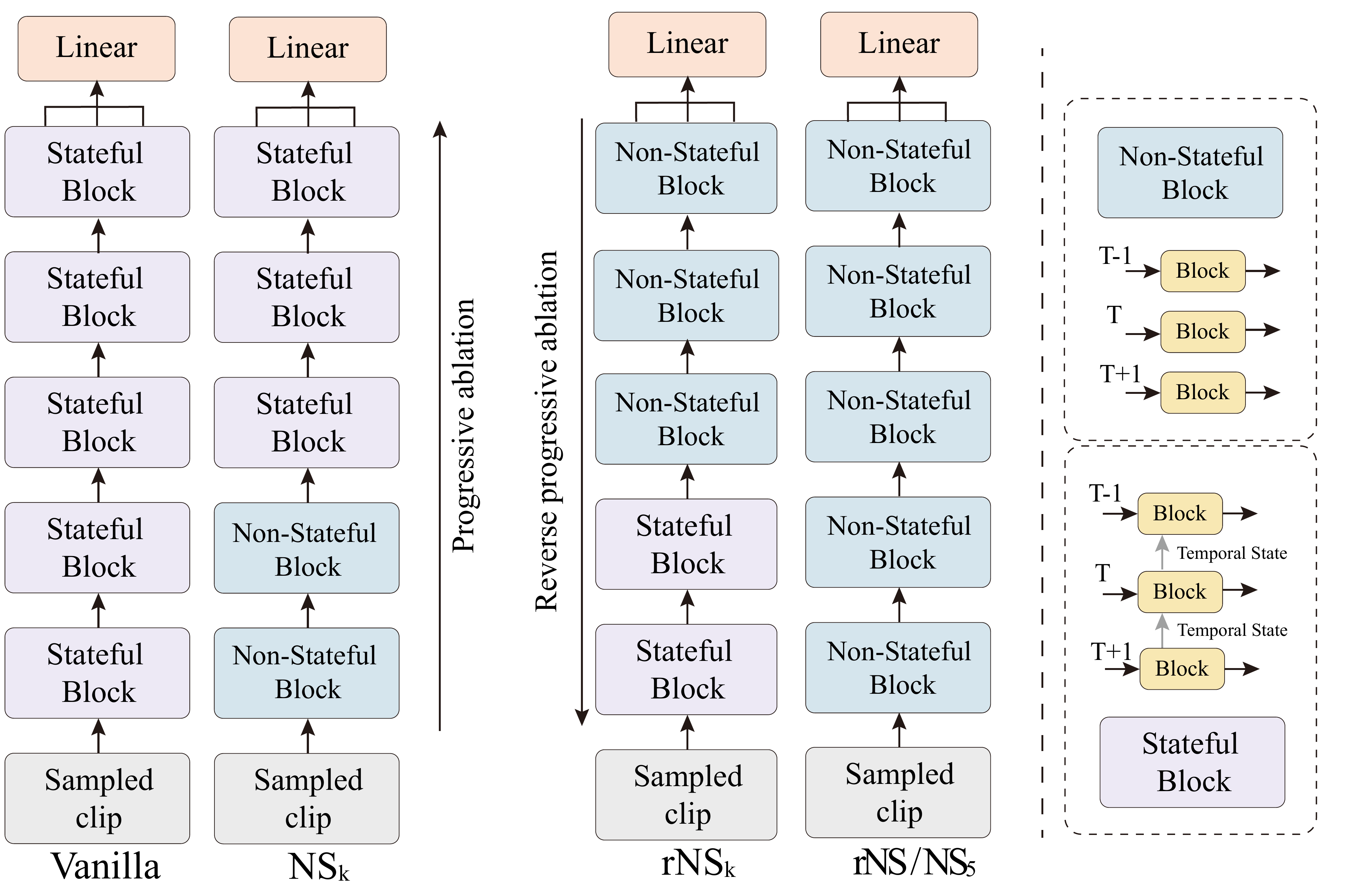}
    \caption{\textbf{Illustration of NS models}, demonstrating forward (\texttt{NS}) and reverse (\texttt{rNS}) strategies for progressive removal of temporal state across layers. }
    \label{fig:NS}
    \vspace{-1.8em}
\end{figure}

In this section, we examine various SNN variants to gain intuitive insights into their temporal modeling capabilities. To investigate the contribution of membrane potential propagation to temporal modeling in SNN, we design Non-Stateful (\textbf{\texttt{NS}}) network variants for ablation analysis (Figure \ref{fig:NS}).

\textbf{Problem Formulation.} Given sampled video input $\mathcal{V} = \{F_1, F_2, ..., F_T\}$ where $F_t \in \mathbb{R}^{3 \times H \times W}$ denotes the $t$-th frame. The SNN adopts a ResNet-like architecture\cite{fang2021deep} with $L=5$ stages $\{\mathcal{H}^1, \mathcal{H}^2, \mathcal{H}^3, \mathcal{H}^4, \mathcal{H}^5\}$, corresponding to the stem layer and four residual stages. For the layer in stage $l$ at time step $t$, we denote the input, membrane potential, and spike output as $I_t^l \in \mathbb{R}^{C_l \times H_l \times W_l}$, $V_t^l \in \mathbb{R}^{C_l \times H_l \times W_l}$, and $S_t^l \in \{0,1\}^{C_l \times H_l \times W_l}$, respectively.

We treat membrane potential $V_t^l$ as the \textbf{internal state} that accumulates temporal information. The output at time $t$ depends on both current input $I_t^l$ and historical state $V_{t-1}^l$:
\begin{equation}
S_t^l = f(I_t^l, V_{t-1}^l; \theta^l)
\end{equation}
We define the standard SNN preserving membrane potential propagation as the \textbf{Stateful Model}, and the variant removing this mechanism as the \textbf{Non-Stateful Model}, whose response degenerates to an instantaneous mapping $\tilde{S}_t^l = f(I_t^l; \theta^l)$. Comparing their performance on temporal tasks, we can quantify the contribution of state propagation to temporal modeling capability.

\textbf{Non-Stateful Variant.} The Leaky Integrate-and-Fire (LIF) \cite{gerstner2002spiking,burkitt2006review} neuron maintains state across time steps:
\begin{equation}
\begin{aligned}
&V_t^l = (1-\frac{1}{\tau}) V_{t-1}^l \odot (1-S_{t-1}^l) + \frac{1}{\tau} I_t^l \\
&S_t^l = \Theta(V_t^l - V_{th}) \\
&I_{t}^{l+1} = \mathcal{F}^{l+1}(S_t^l)
\end{aligned}
\end{equation}
where $\tau$ is the time constant, $\odot$ denotes element-wise multiplication, $\Theta(\cdot)$ is the Heaviside function, $V_{th}$ is the threshold, and $\mathcal{F}^{l+1}(\cdot)$ represents the residual block transformation. The initial state is $V_0^l = \mathbf{0}$. The $V_{t-1}^l$ term enables temporal integration, while the reset mechanism $(1-S_{t-1}^l)$ ensures reset after firing.

In the NS model, we remove membrane potential propagation at specific stages, simplifying the dynamics to:
\begin{equation}
\tilde{V}_t^l = \frac{1}{\tau} I_t^l
\end{equation}
Neurons lose temporal capability, degrading to a stateless mapping independent of historical information $\{V_{t'}^l\}_{t'<t}$.

\textbf{Progressive Ablation Study.} We define two opposite progressive ablation strategies to analyze state propagation importance across network hierarchies:

\textbf{(i) Forward Ablation NS$k$:} Applies Non-Stateful neurons from stage 1 to $k$ ($k\in\{1,2,3,4,5\}$):
\begin{equation}
V_t^l = 
\begin{cases}
\frac{1}{\tau} I_t^l, & l \leq k \\
(1-\frac{1}{\tau})V_{t-1}^l \odot (1-S_{t-1}^l) + \frac{1}{\tau} I_t^l, & l > k
\end{cases}
\end{equation}

\textbf{(ii) Reverse Ablation rNS$k$:} Applies Non-Stateful neurons from stage $L$ to $L-k+1$  ($k\in\{1,2,3,4,5\}$):
\begin{equation}
V_t^l = 
\begin{cases}
(1-\frac{1}{\tau})V_{t-1}^l \odot (1-S_{t-1}^l) + \frac{1}{\tau} I_t^l, & l \leq L-k \\
\frac{1}{\tau} I_t^l, & l > L-k
\end{cases}
\end{equation}
\texttt{NS$0$}/\texttt{rNS$0$} correspond to the full stateful model, while \texttt{NS$5$}/\texttt{rNS$5$} represent the completely stateless variant. This bidirectional design enables analysis of temporal integration between bottom stages ($\mathcal{H}^1, \mathcal{H}^2$) and top stages ($\mathcal{H}^4, \mathcal{H}^5$). \texttt{NS} ablation reveals cumulative effects of bottom-layer states, whereas \texttt{rNS} ablation evaluates the top-layer contributions.

\begin{table}[t] 
\centering

\setlength{\tabcolsep}{2pt}
\renewcommand{\arraystretch}{1.2}

\scalebox{0.85}{

\begin{tabular}{lcc|c}
\hline
\multicolumn{1}{l|}{\textbf{stage}} & \multicolumn{2}{c|}{\textbf{Block Structure}}                                                                                                                                                                                                                             & \textbf{Output Size}                          \\ \hline
\multicolumn{1}{l|}{raw clip}       & \multicolumn{2}{c|}{-}                                                                                                                                                                                                                                                    & ${\color[HTML]{cb0000}T}\times$128$\times$128 \\ \hline
\multicolumn{1}{l|}{$conv_1$}       & \begin{tabular}[c]{@{}c@{}}7 $\times$ 7, 64,\\ stride 2$\times$2, \\ padding 3$\times$3,\end{tabular} & \begin{tabular}[c]{@{}c@{}}7 $\times$ 7, 64/$\textbf{{\color[HTML]{009901}r}}$,\\ stride 2$\times$2, $\color[HTML]{3166FF}·p_1$\\ padding 3$\times$3\end{tabular} & ${\color[HTML]{cb0000}T}\times$128$\times$128 \\ \hline
\multicolumn{1}{l|}{$pool_1$}       & \multicolumn{2}{c|}{\begin{tabular}[c]{@{}c@{}}3 $\times$ 3, max,\\ stride 2$\times$2, \\ padding 1$\times$1\end{tabular}}                                                                                                                                                & ${\color[HTML]{cb0000}T}\times$64$\times$64   \\ \hline
\multicolumn{1}{l|}{$stage_2$}      & $\begin{bmatrix} 3 \times 3, 64 \\ 3 \times 3, 64 \end{bmatrix} \times 2$                             & $\left\{\begin{bmatrix} 3 \times 3, 64/\textbf{{\color[HTML]{009901}r}} \end{bmatrix} \times 2\right\}·{\color[HTML]{3166FF}p_2}$                                 & ${\color[HTML]{cb0000}T}\times$32$\times$32   \\ \hline
\multicolumn{1}{l|}{$stage_3$}      & $\begin{bmatrix} 3 \times 3, 128 \\ 3 \times 3, 128 \end{bmatrix} \times 2$                           & $\left\{\begin{bmatrix} 3 \times 3, 128/\textbf{{\color[HTML]{009901}r}} \end{bmatrix} \times 2\right\}·{\color[HTML]{3166FF}p_3}$                                & ${\color[HTML]{cb0000}T}\times$16$\times$16   \\ \hline
\multicolumn{1}{l|}{$stage_4$}      & $\begin{bmatrix} 3 \times 3, 256 \\ 3 \times 3, 256 \end{bmatrix} \times 2$                           & $\left\{\begin{bmatrix} 3 \times 3, 256/\textbf{{\color[HTML]{009901}r}} \end{bmatrix} \times 2\right\}·{\color[HTML]{3166FF}p_4}$                                & ${\color[HTML]{cb0000}T}\times$8$\times$8     \\ \hline
\multicolumn{1}{l|}{$stage_5$}      & $\begin{bmatrix} 3 \times 3, 512 \\ 3 \times 3, 512 \end{bmatrix} \times 2$                           & $\left\{\begin{bmatrix} 3 \times 3, 512/\textbf{{\color[HTML]{009901}r}} \end{bmatrix} \times 2\right\}·{\color[HTML]{3166FF}p_5}$                                & ${\color[HTML]{cb0000}T}\times$4$\times$4     \\ \hline
\multicolumn{3}{c|}{global pool,  fc, temporal avg}                                                                                                                                                                                                                                                             & \#classes                                     \\ \hline
\end{tabular}

}
\caption{\textbf{STSep architecture.} Each block consists of a spatiotemporal separation module with a spatial branch (left) and temporal branch (right). The flag {\color[HTML]{3166FF}{$p$}} determines whether separation is applied. The {\color[HTML]{009901}{$\textbf{r}$}} controls channel reduction ratio. Convolution kernels are specified as $\{S^2, C\}$ for spatial size $S$ and temporal size $C$.} 
\label{tab:structure}
\vspace{-1.5em}
\end{table}

\subsection{Spatial Temporal Separable Network (STSep)}

The \texttt{NS/rNS} ablation study reveals different dependencies on temporal state across different layers in SNNs (details in the Section \ref{sec_exp:NS}). Surprisingly, we observe that removing temporal state in the first/last 1-2 stages actually improves model performance. We attribute this phenomenon to a \textbf{spatio-temporal resource competition} inherent in standard SNN architectures. SNN's neurons usually simultaneously process spatial patterns and dynamic temporal changes within limited representational capacity. While membrane potential states provide temporal capability, they also occupy the neuron's dynamic range, constraining its ability to represent complex spatial patterns.

Therefore, we propose a novel perspective: decoupling temporal and spatial modeling into independent branches to alleviate representational capacity conflicts. Based on this insight, we introduce the \textbf{Spatial-Temporal Separable Network (STSep)}, which explicitly models temporal variations via temporal differences $\Delta X_t = X_t - X_{t-1}$ and separates residual blocks into a dual-pathway architecture, where the spatial branch extracts spatial semantic features and the temporal branch captures dynamic patterns.


\textbf{(i) Spatial Branch} retains the stateless residual branch for extracting spatial semantic features like \texttt{NS}. Given input features $X_t^l \in \mathbb{R}^{C_l \times H_l \times W_l}$, where $t \in \{1,...,T\}$ denotes the temporal step, the spatial branch is defined as $\mathcal{B}^l$:
\begin{equation}
F_t^{s,l} = \mathcal{B}^l(X_t^l; \Theta_s^l)
\end{equation}
where $\Theta_s^l$ denotes the spatial branch parameters and $F_t^{s,l} \in \mathbb{R}^{C_l \times H_l \times W_l}$ represents the spatial feature representation.

\textbf{(ii) Temporal Branch} explicitly extracts motion information through a differencing mechanism. Define the temporal difference operator $\mathcal{D}$ as:
\begin{equation}
\Delta X_t^l = \mathcal{D}(X_t^l, X_{t-1}^l) = X_t^l - X_{t-1}^l
\end{equation}
where $X_{t-1}^l$ is obtained via cache. $\Delta X_t^l$ captures features dynamics,  encoding a discrete approximation of local motion. The temporal branch forward pass is defined as $\mathcal{T}^l$:
\begin{equation}
F_t^{t,l} = \mathcal{T}^l(\Delta X_t^l; \Theta_t^l)
\end{equation}
where $\Theta_t^l$ denotes the temporal branch parameters and $F_t^{t,l} \in \mathbb{R}^{C_l \times H_l \times W_l}$ represents the temporal feature representation. In particular, $X_0^l = \mathbf{0}_{C_l \times H_l \times W_l}$. 

\textbf{(iii) Spatial-Temporal Separable Block} Outputs from spatial and temporal branches are fused via addition and combined with residual connection\cite{he2016deep}:
\begin{equation}
X_{t}^{l+1} = X_t^l + (1-\alpha^l) F_t^{s,l} + \alpha^l F_t^{t,l}
\end{equation}
where $\alpha^l \in [0,1]$ is the scaling coefficient for all layers, balancing spatial and temporal feature contributions.

For implementation, the temporal transformation $\mathcal{T}^l$ is implemented using a single 3$\times$3 convolutional layer $W_t^l \in \mathbb{R}^{C_l/r \times C_l/r \times 3 \times 3}$. Parameters are initialized by copying the first convolution of the spatial branch to avoid feature mismatch. The scaling factor is uniformly set as $\alpha^l = 0.25$.

From a computational cost perspective, after resolution and channel scaling, the temporal difference operation and the temporal transformation only introduces a few FLOPs.

%% file: sec/4_experiments.tex
\section{Experiment}

\paragraph{Training}
All models are trained end-to-end using the AdamW optimizer \cite{loshchilov2017decoupled}, with a learning rate of $6e^{-4}$ and weight decay of $5e^{-6}$. Training employs a cosine annealing schedule. We maintain a batch size of $256$ and, consistent with \cite{goyal2017accurate}, the learning rate scales linearly with the batchsize.

When leveraging pretrained weights, the temporal difference branch's convolutional weights are initialized by copying the corresponding weights from the spatial branch. Experiments show that \textit{random} or \textit{zero} initialization causes performance degradation and training instability. Input frames are sampled at stride $2$, with sequence lengths of $8/16$ frames. TSN sampling \cite{wang2016temporal} uniformly divides videos into $n$ segments, with randomly sampled one frame within each segment. Spatial resolution is fixed at $128\times128$ with random scaling and cropping. Experiments indicate larger resolutions provide marginal gains at significant computational cost. Data augmentation details are in the Appendix.

\paragraph{Inference}
Because video durations($N$) far exceed typical clip lengths ($N>50$), we follow established practice \cite{wang2016temporal,feichtenhofer2019slowfast} by uniformly sampling $M$ clips (e.g. $M=3$) across the full video and averaging their softmax scores to produce final predictions. In TSN \cite{wang2016temporal}, multiple($M$) temporal samplings are performed, with scores averaged across all clips as well. For SNN, inference matches training stage \cite{wu2019direct,zheng2021going}, averaging outputs over time steps. \texttt{"GFLOPs $\times$ views"} are computed for inference complexity. We use equivalent \texttt{FLOPs} for floating operations on GPUs as computational cost metric for convenient comparison, while dedicated neuromorphic hardware would achieve lower computation.

\paragraph{Datasets}
The Something-Something V2 (SSV2) dataset \cite{goyal2017something} is a large-scale benchmark for temporally-dependent action recognition, comprising $220,847$ videos across $174$ fine-grained classes collected via crowdsourcing. Its design enforces reliance on temporal dynamics rather than static appearance, making it a standard testbed for temporal representation learning. Unlike earlier datasets such as UCF101 and HMDB51 \cite{xie2018rethinking,zhou2018temporal}, which exhibit strong static bias allowing single-frame cues to rival temporal models, SSV2 removes such shortcuts, ensuring temporal modeling is decisive for performance. We use SSV2 as our primary benchmark and report generalization on UCF101 \cite{soomro2012ucf101} and HMDB51 \cite{kuehne2011hmdb}, which containing $13,320/6,766$ videos in $101/51$ classes across domains of human actions. All experiments are evaluated on split \texttt{1} for reproducibility.

\begin{figure}[t]
    \centering
    \includegraphics[width=0.95\linewidth]{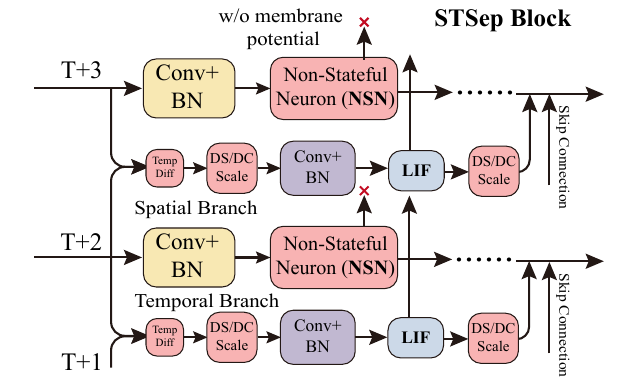}
    \vspace{-0.5em}
    \caption{\textbf{Detailed illustration of the STSep module}, where \texttt{NSN} denotes Non-Stateful Neuron. \texttt{DS/DC scale} represents downsampling and channel scaling operations.}
    \label{fig:STSep_expand}
    \vspace{-1.8em}
\end{figure}

\begin{table*}[] 

\scalebox{0.9}{ 
\begin{subtable}[t]{0.4\textwidth}
\centering
\setlength{\tabcolsep}{2pt}
\renewcommand{\arraystretch}{1.2}

\begin{tabular}{l|cccc|cc}
\multicolumn{1}{c|}{}   & \multicolumn{4}{c|}{\textbf{8$\times$128}}                                                          & \multicolumn{2}{c}{\textbf{16$\times$128}}    \\ \hline
\textbf{input}          & \multicolumn{2}{c|}{{\ul \textbf{$\tau = 1$}}}     & \multicolumn{2}{c|}{{\ul \textbf{$\tau = 2$}}} & \multicolumn{2}{c}{{\ul \textbf{$\tau = 1$}}} \\ \cline{1-1}
\textbf{model}          & \textbf{Top1} & \multicolumn{1}{c|}{\textbf{Top5}} & \textbf{Top1}          & \textbf{Top5}         & \textbf{Top1}          & \textbf{Top5}        \\ \hline
\textbf{vanilla}        & 21.1          & \multicolumn{1}{c|}{45.2}          & 21.1                   & 45.2                  & 24.6                   & 50.7                 \\
\hdashline \textbf{NS1} & 22.6          & \multicolumn{1}{c|}{48.3}          & 21.9                   & 46.4                  & 24.8                   & 51.8                 \\
\textbf{NS2}            & 22.7          & \multicolumn{1}{c|}{48.5}          & 23.5                   & 49.1                  & \textbf{26.6}          & \textbf{53.3}        \\
\textbf{NS3}            & 22.7          & \multicolumn{1}{c|}{47.8}          & 24.1                   & 50.1                  & 25.5                   & 51.8                 \\
\textbf{NS4}            & 19.6          & \multicolumn{1}{c|}{42.9}          & 21.9                   & 46.8                  & 19.6                   & 43.1                 \\
\textbf{NS5/rNS5}       & 10.3          & \multicolumn{1}{c|}{28.0}          & 12.4                   & 32.5                  & 10.1                   & 28.2                 \\
\textbf{rNS4}           & 17.8          & \multicolumn{1}{c|}{40.1}          & 18.2                   & 41.7                  & 17.2                   & 39.9                 \\
\textbf{rNS3}           & \textbf{24.0} & \multicolumn{1}{c|}{\textbf{49.2}} & \textbf{25.6}          & \textbf{52.2}         & 24.5                   & 50.0                 \\
\textbf{rNS2}           & 22.9          & \multicolumn{1}{c|}{47.9}          & {\ul 25.6}             & {\ul 51.1}            & {\ul 25.5}             & {\ul 51.8}           \\
\textbf{rNS1}           & {\ul 23.9}    & \multicolumn{1}{c|}{{\ul 48.8}}    & 24.6                   & 49.6                  & 25.3                   & 51.6                
\end{tabular}

\caption{\textbf{Comparison with NS/rNS model on Something-Something V2.} Table present \texttt{NS/rNS} model performance on SSV2 across time constants $\tau\in\{1,2\}$ and time steps $T\in\{8,16\}$. Moderate state information removal improves performance over the vanilla baseline, while excessive state removal causes severe degradation.}
\label{tab:NS}
\end{subtable}
}
\scalebox{0.9}{ 
\begin{subtable}[t]{0.32\textwidth}
\renewcommand{\arraystretch}{1.4}
\setlength{\tabcolsep}{1.5pt}
\centering

\begin{tabular}{l|c|c|cc}
\textbf{stage}              & \textbf{params} & \textbf{FLOPs} & \textbf{Top1}               & \textbf{Top5}               \\ \hline
\textbf{vanilla}            & 11.3M           & 9.48G          & 24.6                        & 50.7                        \\
\hdashline \textbf{stage 1} & 11.3M           & 9.58G          & {\color[HTML]{9A0000} 28.5} & {\color[HTML]{9A0000} 56.0} \\
\textbf{stage 1-2}          & 11.3M           & 9.69G          & {\color[HTML]{9A0000} 25.8} & {\color[HTML]{9A0000} 52.2} \\
\textbf{stage 1-3}          & 11.4M           & 9.78G          & {\color[HTML]{9A0000} 26.6} & {\color[HTML]{9A0000} 54.6} \\
\textbf{stage 1-4}          & 11.6M           & 9.87G          & {\color[HTML]{036400} 32.5} & {\color[HTML]{036400} 60.8} \\
\textbf{stage 1-5}          & 11.8M           & 9.89G          & {\color[HTML]{036400} 34.9} & {\color[HTML]{036400} 63.7} \\
\textbf{stage 2-5}          & 11.8M           & 9.79G          & {\color[HTML]{036400} 30.1} & {\color[HTML]{036400} 57.9} \\
\textbf{stage 3-5}          & 11.8M           & 9.68G          & {\color[HTML]{9A0000} 27.9} & {\color[HTML]{9A0000} 54.5} \\
\textbf{stage 4-5}          & 11.5M           & 9.59G          & {\color[HTML]{036400} 29.7} & {\color[HTML]{036400} 56.6} \\
\textbf{stage 5}            & 11.5M           & 9.50G          & {\color[HTML]{9A0000} 27.7} & {\color[HTML]{9A0000} 54.5}
\end{tabular}

\caption{\textbf{Comparison with STSep in different stage on Something-Something V2.} The table shows STSep performance with progressive spatiotemporal separation across stages. \textbf{\texttt{\textcolor[HTML]{036400}{Green}}} indicates improvements over \texttt{Average}, while \textbf{\texttt{\textcolor[HTML]{9A0000}{red}}} denotes degradation.}
\label{tab:stage}
\end{subtable}
}
\scalebox{0.85}{
\begin{subtable}[t]{0.4\textwidth}
\renewcommand{\arraystretch}{1.2}
\setlength{\tabcolsep}{1.2pt}
\centering

\begin{tabular}{r|rcc|cc}
\multicolumn{1}{l|}{} & \multicolumn{1}{c}{\textbf{setting}}                        & \multicolumn{1}{l}{\textbf{params}} & \multicolumn{1}{l|}{\textbf{FLOPs}} & \multicolumn{1}{l}{\textbf{Top1}} & \multicolumn{1}{l}{\textbf{Top5}} \\ \hline
STSep                 & -                                                           & 11.5M                               & 9.60G                               & 33.8                              & 62.9                              \\ \hline
STSep                 & w/o diff                                                    & 11.5M                               & 9.60G                               & 25.5                              & 52.3                              \\
STSep                 & w/o conv                                                    & 11.3M                               & 9.48G                               & 26.8                              & 54.6                              \\
\hdashline STSep      & \begin{tabular}[c]{@{}r@{}}w/o spatial\\ brach\end{tabular} & 3.26M                               & 7.39G                               & 19.6                              & 40.1                              \\
\hdashline SE block   & -                                                           & 11.4M                               & 9.60G                               & 26.3                              & 51.8                              \\ \hline
STSep                 & r = 16                                                      & 11.5M                               & 9.60G                               & 31.1                              & 58.8                              \\
STSep                 & 8                                                           & 11.8M                               & 9.72G                               & 31.6                              & 59.1                              \\
STSep                 & 4                                                           & 12.3M                               & 9.96G                               & 32.1                              & 59.1                              \\
STSep                 & 2                                                           & 13.3M                               & 10.4G                               & 32.7                              & 60.4                              \\
\hdashline STSep      & r = 1/ s=1                                                  & 15.4M                               & 11.4G                               & 33.6                              & 61.6                              \\
\hdashline STSep      & s = 2                                                       & 15.4M                               & 10.3G                               & 33.8                              & 62.9                              \\
STSep                 & 4                                                           & 15.4M                               & 10.1G                               & 32.7                              & 60.7                             
\end{tabular}

\caption{\textbf{Ablation Study of STSep.} The table presents STSep performance under different architectural variants and hyperparameters, including 1) removal of temporal difference, 2) convolution in temporal branch, or 3) spatial branches, 4) replacement with \texttt{SE block}, and  variations in 5) channel reduction ratio \texttt{$r$} and 6) spatial downsampling factor \texttt{$s$}.}
\label{tab:ablation}
\end{subtable}
}
\vspace{-0.5em}
\caption{The table presents three sets of experiments: a) \texttt{NS/rNS} model on SSV2 across varying time constants $\tau$ and timesteps $T$; b) STSep with progressively stage-wise spatiotemporal separation; and c) STSep ablation studies across component configurations and hyperparameters. All experiments employ SEW-ResNet18\cite{fang2021deep} as the backbone, trained from scratch for 50 epochs. Frames resolution adopts $128\times128$ with TSN sampling strategy. Evaluation employs 3-clip testing, reporting both \texttt{Top-1} and \texttt{Top-5} on the validation set.}
\label{tab:analysis}
\vspace{-1.5em}
\end{table*}

\subsection{Non-stateful Model}
\label{sec_exp:NS}
Table~\ref{tab:NS} presents the performance of Non-Stateful (\texttt{\textbf{NS}}) and reverse Non-Stateful (\texttt{\textbf{rNS}}) models on Something-Something V2\cite{goyal2017something}. Experiments vary temporal length $T \in \{8, 16\}$, spatial resolution $128\times128$, and time constant $\tau \in \{1, 2\}$. All models adopt SEW-ResNet18\cite{fang2021deep} backbone, with \texttt{"vanilla"} denoting the standard SNN baseline preserving membrane potential across all stages.

Several key observations emerge from the results:

\textbf{(i) Removing membrane potential propagation does not always degrade performance} Under moderate state removal (\texttt{NS1/NS2} and \texttt{rNS2/rNS3}), models surprisingly outperform the vanilla baseline. For instance, under the $\tau=2, T=8$, both \texttt{NS2} and \texttt{rNS3} surpass \texttt{vanilla}. However, excessive removal leads to severe degradation. \texttt{NS5/rNS5} exhibit catastrophic collapse across all configurations, with Top-1 accuracy dropping to $10-12\%$.  This non-monotonic pattern persists across all three experimental settings and generalizes to other datasets (UCF101\cite{soomro2012ucf101}, HMDB51\cite{kuehne2011hmdb}, in Appendix), indicating its generality rather than coincidence.

We attribute the performance gains to resource competition in spatial-temporal modeling. LIF neurons need encode both spatial semantics and temporal dynamics within limited dynamic range. Membrane potential maintenance, while enabling temporal integration, consumes representational resources that could otherwise be allocated to learning complex spatial semantics. Moderate state removal liberates these resources, enabling more effective extraction of spatial features. However, excessive removal eliminates essential temporal modeling capacity, preventing dynamic pattern capture in videos and causing performance collapse.

\textbf{(ii) Forward(\texttt{NS}) and reverse(\texttt{rNS}) ablation exhibit asymmetric performance patterns.} Removing states from top layers (\texttt{rNS}) impacts performance more dramatically than removing from bottom layers (\texttt{NS}), and removal from bottom layers has a more pronounced effect(See Figure \ref{fig:one_stage}(a)). For example, both \texttt{Stage4} and \texttt{Stage5} outperform \texttt{stage1} and \texttt{Stage2}. Also, under $\tau=2, T=8$, \texttt{rNS4} degrades faster than \texttt{NS4}. Similar phenomena are observed across all configurations in the table.

We hypothesize that top-layer blocks primarily process high-level semantic features, where temporal correlations significantly compete for action recognition resources, while bottom-layer blocks extract low-level spatial features (edges, textures) with relatively minor impact. When the network need maintain a certain number of temporal states, releasing top state could be relatively more beneficial. ANN-based models\cite{tran2018closer} is also align with the importance of top-layer.

These findings yield two key insights for SNN architecture design: (i) Spatial and temporal information should be separated to avoid resource competition. (ii) Not all layers require state information, and different temporal strategies should be adopted across depths.

\subsection{ Spatial-Temporal Separable Network (STSep)}
In this section, we investigate the influence of explicit spatiotemporal separation in spiking neural networks.  A question arises: \textit{  does STSep's spatiotemporal separation exhibit similar layer-wise dependencies as observed in \texttt{NS}? can the design insights from \texttt{NS} inform STSep configuration?}

We conduct similar ablation studies, progressively applying spatiotemporal separation to blocks at different stages. Experiments adopt the same configuration as \texttt{NS}: $T=16$ frames, $128\times128$ resolution, and $\tau=1$. Results are presented in Table~\ref{tab:stage}. The experiments reveal a striking phenomenon: \textit{STSep exhibits a complementary performance pattern to \texttt{NS}.} Spatiotemporal separation in bottom and top layers significantly improves performance, corresponding to the stages where removing temporal state in \texttt{NS} yielded gains. Notably, applying STSep at stage1 and stage5 produces the most pronounced impact, while separation in intermediate layers shows more moderate effects. This lends support to our hypothesis: temporal state may cause resource conflicts that impair spatial semantic processing. While \texttt{NS} trades off temporal capacity for spatial capacity, STSep achieves synergistic enhancement of both via spatiotemporal separation.

\begin{figure}[t]
    \centering
    \includegraphics[width=1.05\linewidth]{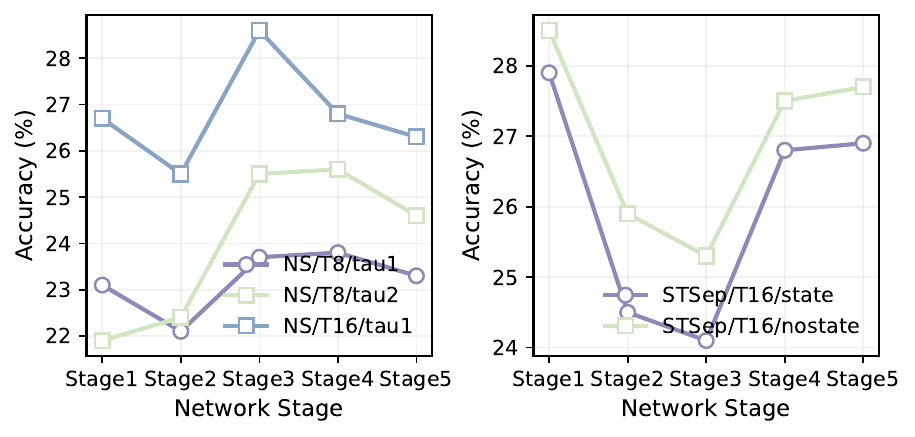}
    \vspace{-1.8em}
    \caption{\textbf{Comparison of stage-wise ablation.} The x-axis represents different stages of separation, while the y-axis shows \texttt{Top-1} accuracy. a) \texttt{NS/rNS} model comparison with various configurations. b) \texttt{STSep} model comparison in \texttt{state} vs. \texttt{nostate}.}
    \label{fig:one_stage}
    \vspace{-1.8em}
\end{figure}

Beyond progressive ablation, we identify the marginal contribution of applying STSep at each individual stage. Figure~\ref{fig:one_stage}(b) presents results when STSep is applied to a single stage only. This pattern likely reflects hierarchical specialization that shallow and deep layers, which handle feature encoding and semantic extraction respectively, both demand substantial spatial capacity, thus benefiting most from spatiotemporal separation.

Given efficiency constraints, we need balance performance gains against computational overhead. Unlike 3D Convs in ANNs that progressively downsample temporal dimensions, SNNs process the complete temporal sequence throughout all layers. Moreover, deeper layers possess substantially larger channel counts. Therefore, we selectively apply separation at stages $k \in \{1, 2, 5\}$, preserving strong performance while controlling computational overhead.

\begin{table}[t] 
\centering

\setlength{\tabcolsep}{3pt}
\renewcommand{\arraystretch}{1.2}
\scalebox{0.87}{

\begin{tabular}{l|c|c|c|c|cc}
\textbf{Method}                              & \multicolumn{1}{l|}{\textbf{Pre}}         & \textbf{Params} & \multicolumn{1}{l|}{\textbf{FLOPs$\times$views}} & \multicolumn{1}{l|}{\textbf{Frames}} & \multicolumn{1}{l}{\textbf{Top1}} & \multicolumn{1}{l}{\textbf{Top5}} \\ \hline
TSN \cite{wang2018temporal}                  & -                                         & 11.3M           & 9.48 $\times$ 3 G                                & 16$\times$128                        & 24.9                              & 51.8                              \\
\hdashline PLIF \cite{fang2021incorporating} & \multirow{6}{*}{$\rotatebox{90}{ImgNet}$} & 11.3M           & 9.48 $\times$ 3 G                                & 16$\times$128                        & 26.5                              & 53.5                              \\
TET \cite{deng2022temporal}                  &                                           & 11.3M           & 9.48 $\times$ 3 G                                & 16$\times$128                        & 22.1                              & 47.2                              \\
RSNN \cite{zhang2019spike}                   &                                           & 11.3M           & 9.48 $\times$ 3 G                                & 16$\times$128                        & 26.4                              & 53.1                              \\
TDBN \cite{zheng2021going}                   &                                           & 11.3M           & 9.48 $\times$ 3 G                                & 16$\times$128                        & 27.4                              & 55.2                              \\
Tcja \cite{zhu2024tcja}                      &                                           & 11.3M           & 9.49 $\times$ 3 G                                & 16$\times$128                        & 24.1                              & 49.1                              \\
TKS \cite{dong2024temporal}                  &                                           & 11.3M           & 9.48 $\times$ 3 G                                & 16$\times$128                        & 24.5                              & 49.4                              \\ \hline
\textbf{STSep}                               & \multirow{3}{*}{-}                        & 11.5M           & 9.60 $\times$ 3 G                                & 8$\times$128                         & 26.5                              & 52.4                              \\
\textbf{STSep}                               &                                           & 11.5M           & 9.60 $\times$ 10 G                               & 8$\times$128                         & 28.7                              & 55.8                              \\
\textbf{STSep+TSN}                           &                                           & 11.5M           & 9.60 $\times$ 3 G                                & 8$\times$128                         & 29.8                              & 58.1                              \\
\hdashline \textbf{STSep}                    & \multirow{2}{*}{-}                        & 11.5M           & 9.60 $\times$ 3 G                                & 16$\times$128                        & 32.9                              & 61.3                              \\
\textbf{STSep}                               &                                           & 11.5M           & 9.60 $\times$ 10 G                               & 16$\times$128                        & 33.3                              & 61.9                              \\ \hline
\textbf{STSep(S1-2)}                         & \multirow{3}{*}{$\rotatebox{90}{ImgNet}$} & 11.3M           & 9.58 $\times$ 3 G                                & 16$\times$128                        & \textbf{28.5}                     & \textbf{56.0}                     \\
\textbf{STSep}                               &                                           & 11.5M           & 9.60 $\times$ 3 G                                & 16$\times$128                        & \textbf{33.7}                     & \textbf{62.5}                     \\
\textbf{STSep}                               &                                           & 11.5M           & 9.60 $\times$ 10 G                               & 16$\times$128                        & \textbf{34.4}                     & \textbf{62.8}                    
\end{tabular}

}
\caption{\textbf{Comparison with SNN temporal modeling methods on Something-Something V2.} \textbf{Bold} indicates best performance, \underline{underline} indicates second-best. \texttt{"Pre"} denotes if ImageNet pretraining are used. \texttt{"S1-2"} indicates STSep applied at stages 1,2.}
\label{tab:SSV2}
\vspace{-1.5em}
\end{table}

\subsection{Ablation Experiments}
\vspace{-0.5em}

\begin{table}[t] 
\centering

\setlength{\tabcolsep}{3pt}
\renewcommand{\arraystretch}{1.2}
\scalebox{0.87}{

\begin{tabular}{l|c|c|c|c|cc}
\textbf{Method}                              & \multicolumn{1}{l|}{\textbf{Pre}}         & \textbf{Params} & \multicolumn{1}{l|}{\textbf{FLOPs$\times$views}} & \multicolumn{1}{l|}{\textbf{Frames}} & \multicolumn{1}{l}{\textbf{Top1}} & \multicolumn{1}{l}{\textbf{Top5}} \\ \hline
TSN \cite{wang2018temporal}                  & -                                         & 11.3M           & 9.48 $\times$ 3 G                                & 16$\times$128                        & 43.3                              & 70.8                              \\
\hdashline PLIF \cite{fang2021incorporating} & \multirow{6}{*}{$\rotatebox{90}{ImgNet}$} & 11.3M           & 9.48 $\times$ 3 G                                & 16$\times$128                        & 66.9                              & 88.5                              \\
TET \cite{deng2022temporal}                  &                                           & 11.3M           & 9.48 $\times$ 3 G                                & 16$\times$128                        & 67.4                              & 90.0                              \\
RSNN \cite{zhang2019spike}                   &                                           & 11.3M           & 9.48 $\times$ 3 G                                & 16$\times$128                        & 66.6                              & 88.4                              \\
TDBN \cite{zheng2021going}                   &                                           & 11.3M           & 9.48 $\times$ 3 G                                & 16$\times$128                        & 68.1                              & 90.2                              \\
Tcja \cite{zhu2024tcja}                      &                                           & 11.3M           & 9.49 $\times$ 3 G                                & 16$\times$128                        & 67.2                              & 89.8                              \\
TKS \cite{dong2024temporal}                  &                                           & 11.3M           & 9.48 $\times$ 3 G                                & 16$\times$128                        & 67.6                              & 90.0                              \\ \hline
\textbf{STSep}                               & \multirow{2}{*}{-}                        & 11.5M           & 9.60 $\times$ 1 G                                & 8$\times$128                         & 43.5                              & 69.9                              \\
\textbf{STSep}                               &                                           & 11.5M           & 9.60 $\times$ 3 G                                & 8$\times$128                         & 46.7                              & 73.3                              \\
\hdashline \textbf{STSep}                    & \multirow{2}{*}{-}                        & 11.5M           & 9.60 $\times$ 1 G                                & 16$\times$128                        & 47.9                              & 74.2                              \\
\textbf{STSep}                               &                                           & 11.5M           & 9.60 $\times$ 3 G                                & 16$\times$128                        & 48.5                              & 74.7                              \\ \hline
\textbf{STSep}                               & \multirow{3}{*}{$\rotatebox{90}{ImgNet}$} & 11.5M           & 9.60 $\times$ 1 G                                & 16$\times$128                        & \textbf{65.7}                     & \textbf{88.8}                     \\
\textbf{STSep(S1-2)}                         &                                           & 11.3M           & 9.58 $\times$ 3 G                                & 16$\times$128                        & \textbf{68.7}                     & \textbf{90.3}                     \\
\textbf{STSep}                               &                                           & 11.5M           & 9.60 $\times$ 3 G                                & 16$\times$128                        & \textbf{69.5}                     & \textbf{90.7}                    
\end{tabular}

}
\caption{\textbf{Comparison with SNN temporal modeling methods on UCF101.} STSep maintains superior performance on UCF101 despite the dataset's heavier reliance on static scene cues. similarly, \textbf{Bold} indicates best performance, \underline{underline} indicates second-best. \texttt{"Pre"} denotes whether ImageNet pretrained weights are used.}
\label{tab:ucf101}
\vspace{-1.5em}
\end{table}

\paragraph{w/o diff}
The preceding experiments have validated the effectiveness of STSep. We further investigate whether the observed improvements stem from temporal difference modeling or merely from architectural modifications. We replace the temporal branch's input from difference features $\Delta X_t = X_t - X_{t-1}$ to current features $X_t$, yielding the \texttt{w/o diff} variant. As shown in Table~\ref{tab:ablation}, this variant substantially degrades performance while keeping structure and parameters unchanged, confirming that temporal difference provides critical motion cues to temporal branch, rather than performance gains arising from architectural redundancy.
 
\paragraph{w/o Conv}
We further examine whether temporal difference can be directly incorporated, similar to residual connections, without convolutional transformations. We remove all convolutions from the temporal branch while retaining only the difference operation. For blocks requiring downsampling, we use $1\times 1$ convolutions to adjust spatial resolution and channel dimensions, consistent with residual connection (adding negligible parameters). As shown in Table~\ref{tab:ablation}, \texttt{w/o conv} degrades in performance but still surpasses the vanilla baseline, showing that explicit temporal differences aid convergence, while convolutional transformations further refine representations to capture sophisticated motion patterns.

\paragraph{w/o spatial branch}
We investigate the temporal branch contribution in isolation by removing spatial branches from separated stages while preserving non-separated stages. Results show that retaining only the temporal branch fails to maintain original performance, despite the temporal branch providing modest improvements with fewer parameters. This underscores the complementary nature of spatial semantic and temporal dynamics, where neither information source alone sufficiently supports discriminative capacity.

\begin{table}[t] 
\centering

\setlength{\tabcolsep}{3pt}
\renewcommand{\arraystretch}{1.2}
\scalebox{0.87}{

\begin{tabular}{l|c|c|c|c|cc}
\textbf{Method}                              & \multicolumn{1}{l|}{\textbf{Pre}}         & \textbf{Params} & \multicolumn{1}{l|}{\textbf{FLOPs$\times$views}} & \multicolumn{1}{l|}{\textbf{Frames}} & \multicolumn{1}{l}{\textbf{Top1}} & \multicolumn{1}{l}{\textbf{Top5}} \\ \hline
TSN \cite{wang2018temporal}                  & -                                         & 11.3M           & 9.48 $\times$ 3 G                                & 16$\times$128                        & 17.2                              & 48.2                              \\
\hdashline PLIF \cite{fang2021incorporating} & \multirow{6}{*}{$\rotatebox{90}{ImgNet}$} & 11.3M           & 9.48 $\times$ 3 G                                & 16$\times$128                        & 39.1                              & 73.2                              \\
TET \cite{deng2022temporal}                  &                                           & 11.3M           & 9.48 $\times$ 3 G                                & 16$\times$128                        & 39.6                              & 74.0                              \\
RSNN \cite{zhang2019spike}                   &                                           & 11.3M           & 9.48 $\times$ 3 G                                & 16$\times$128                        & 39.9                              & 73.8                              \\
TDBN \cite{zheng2021going}                   &                                           & 11.3M           & 9.48 $\times$ 3 G                                & 16$\times$128                        & 40.1                              & 73.9                              \\
Tcja \cite{zhu2024tcja}                      &                                           & 11.3M           & 9.49 $\times$ 3 G                                & 16$\times$128                        & 39.9                              & 73.5                              \\
TKS \cite{dong2024temporal}                  &                                           & 11.3M           & 9.48 $\times$ 3 G                                & 16$\times$128                        & 40.0                              & 74.0                              \\ \hline
TET \cite{deng2022temporal}                  & \multirow{2}{*}{-}                        & 11.3M           & 9.48 $\times$ 3 G                                & 16$\times$128                        & 18.7                              & 47.2                              \\
TDBN \cite{zheng2021going}                   &                                           & 11.3M           & 9.48 $\times$ 3 G                                & 16$\times$128                        & 19.1                              & 49.5                              \\ \hline
\textbf{STSep}                               & -                                         & 11.5M           & 9.60 $\times$ 3 G                                & 8$\times$128                         & 20.8                              & 50.9                              \\
\hdashline \textbf{STSep}                    & -                                         & 11.5M           & 9.60 $\times$ 3 G                                & 16$\times$128                        & 21.2                              & 50.9                              \\ \hline
\textbf{STSep}                               & \multirow{3}{*}{$\rotatebox{90}{ImgNet}$} & 11.5M           & 9.60 $\times$ 1 G                                & 16$\times$128                        & \textbf{39.8}                     & \textbf{72.4}                     \\
\textbf{STSep(S1-2)}                         &                                           & 11.3M           & 9.58 $\times$ 3 G                                & 16$\times$128                        & \textbf{40.5}                     & \textbf{74.3}                     \\
\textbf{STSep}                               &                                           & 11.5M           & 9.60 $\times$ 3 G                                & 16$\times$128                        & \textbf{41.4}                     & \textbf{74.7}                    
\end{tabular}

}
\caption{\textbf{Comparison with SNN temporal modeling methods on HMDB51.} HMDB51 exhibits similar trends to UCF101 due to their comparable dataset characteristics. }
\label{tab:hmdb51}
\vspace{-1.em}
\end{table}

\begin{figure}[t]
    \centering
    \includegraphics[width=1.0\linewidth]{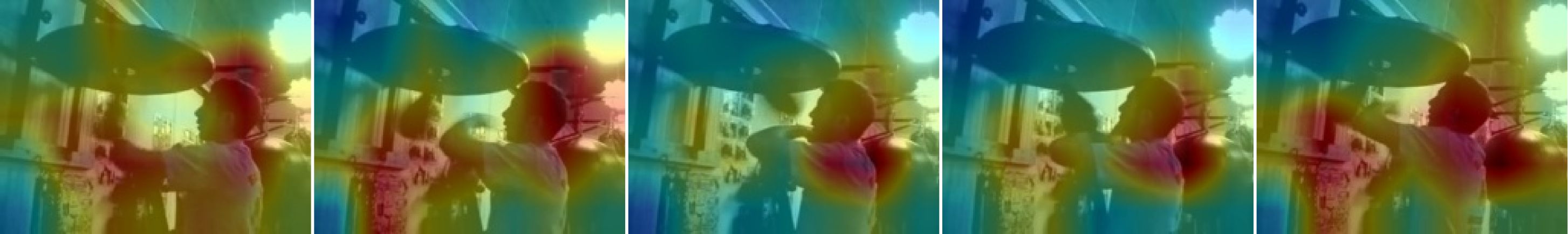}
    \includegraphics[width=1.0\linewidth]{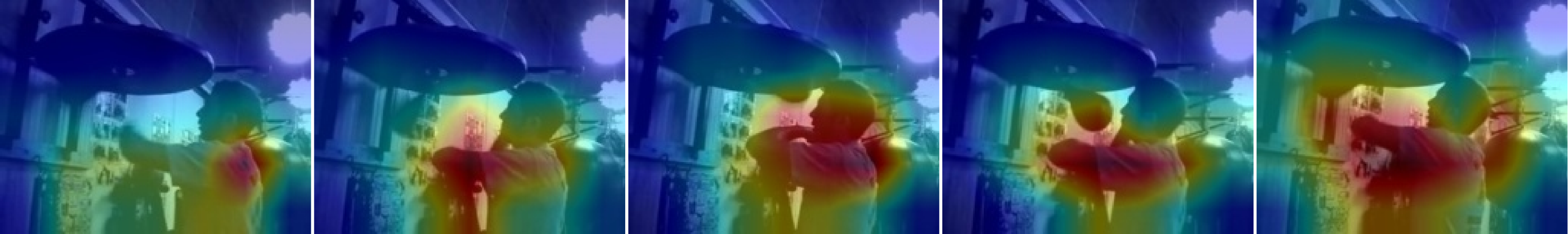}
    \vspace{-1.5em}
    \caption{\textbf{Activation visualization.} \textbf{Top} shows vanilla model activation for sample inputs. \textbf{Bottom} shows STSep model activation for the same samples.}
    \label{fig:heatmap}
    \vspace{-1.8em}
\end{figure}

\begin{table*}[t] 
\centering

\setlength{\tabcolsep}{3.5pt}
\renewcommand{\arraystretch}{1.2}
\scalebox{0.9}{

\begin{tabular}{lccccccccccccccccc}
\hline
                        & \multicolumn{5}{c}{{\ul \textbf{Something-Somethingv2}}}                           & \textbf{} & \multicolumn{5}{c}{{\ul \textbf{UCF101}}}                                          & \textbf{} & \multicolumn{5}{c}{{\ul \textbf{HMDB51}}}                                          \\
\textbf{method}         & \textit{R@3}   & \textit{R@5}   & \textit{R@10}  & \textit{R@20}  & \textit{R@50}  & \textit{} & \textit{R@1}   & \textit{R@3}   & \textit{R@5}   & \textit{R@10}  & \textit{R@20}  & \textit{} & \textit{R@1}   & \textit{R@3}   & \textit{R@5}   & \textit{R@10}  & \textit{R@20}  \\ \hline
\textbf{vanilla}        & 29.74          & 36.93          & 47.32          & 58.39          & 72.65          &           & 40.13          & 48.64          & 53.03          & 59.50          & 67.14          &           & 14.71          & 24.05          & 29.54          & 37.39          & 46.47          \\
\textbf{vanilla+ImgNet} & 30.44          & 37.88          & 47.97          & 58.53          & 72.20          &           & 64.42          & 72.16          & 75.16          & 78.83          & 83.85          &           & 28.95          & 42.75          & 50.52          & 61.44          & 73.01          \\
\textbf{STSep}          & \textbf{38.50} & \textbf{46.50} & \textbf{57.61} & \textbf{67.93} & \textbf{80.18} &           & 41.50          & 49.93          & 55.11          & 61.64          & 69.44          &           & 16.34          & 25.62          & 31.50          & 43.01          & 54.31          \\
\textbf{STSep+ImgNet}   & 36.95          & 44.79          & 55.75          & 66.15          & 78.84          & \textbf{} & \textbf{65.27} & \textbf{73.43} & \textbf{76.90} & \textbf{80.73} & \textbf{85.12} & \textbf{} & \textbf{33.66} & \textbf{48.24} & \textbf{55.10} & \textbf{65.75} & \textbf{75.23}
\end{tabular}

}
\vspace{-0.5em}
\caption{\textbf{STSep performance on video retrieval tasks evaluated on SSV2, UCF101, and HMDB51.} \texttt{R@K} denotes the recall rate when correct samples appear within the \texttt{top-K} retrievals. \textbf{\texttt{+ImgNet}} indicates the ImageNet pretraining.}
\label{tab:retrieval}
\vspace{-1.em}
\end{table*}

\begin{figure*}[t]
    \centering 
    \includegraphics[width=0.99\linewidth]{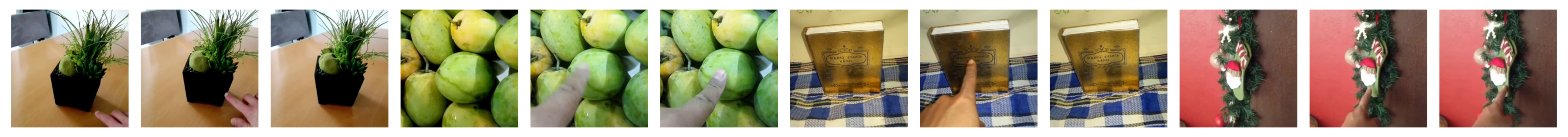}
    \includegraphics[width=0.99\linewidth]{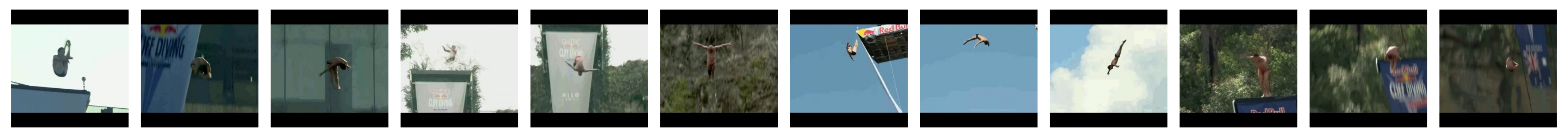}
    \caption{\textbf{Example visualizations of video retrieval.} Each query video sampled from Something-Something V2(Top) and UCF101(Bottom) is shown with its \texttt{top-3} nearest neighbors in feature space, with class labels annotated above each retrieved result. We extract features from the global average pooling layer before classifier, averaging across the temporal dimension for retrieval. }
    \label{fig:retrieval}
    \vspace{-1.5em}
\end{figure*}

\paragraph{vs. SE block}
Furthermore, since SE blocks enhance features through channel-wise recalibration with minimal parameter overhead, we compare this feature enhancement strategies. We replace spatiotemporal separation in STSep with SE blocks. Results show that STSep outperforms SE blocks, demonstrating that explicit spatiotemporal separation is more effective than mixed feature for temporal modeling.

\paragraph{Conv configuration}
The temporal branch convolution size directly impacts model capacity and parameter count. We explore temporal convolution configurations to balance performance and computational cost through channel reduction and spatial downsampling. In Table~\ref{tab:ablation}, $r$ denotes the channel reduction ratio and $s$ denotes the spatial downsampling ratio. Results indicate that excessive reduction significantly degrades performance. We adopt $s=2$ and $r=4$ as the default configuration, effectively reducing computation while maintaining competitive performance.

\paragraph{Activation Visualization}
Finally, We compare spatial attention patterns between STSep and vanilla models. Figure~\ref{fig:heatmap} shows activation visualizations on UCF101 samples (containing object appearance and action information). Vanilla activations concentrate on object appearance and scenes, including humans and equipment, while STSep attends to action-relevant locations and motion trajectories, indicating enhanced sensitivity to dynamic features.

\subsection{Main Results}
Given the scarcity of SNN research on video data, we reproduce several methods that process or enhance temporal features for comparison with STSep~\cite{fang2021incorporating,deng2022temporal,zhang2019spike,zheng2021going,zhu2024tcja,dong2024temporal}. These works approach temporal modeling from diverse perspectives: i) Membrane Potential~\cite{fang2021incorporating}, ii) Recurrent SNNs~\cite{zhang2019spike}, iii) Training Strategy~\cite{deng2022temporal,dong2024temporal}, iv) Feature Interaction~\cite{zhu2024tcja,zheng2021going}.

Table~\ref{tab:SSV2} compares STSep with other temporal modeling approaches for SNNs in SSV2\cite{goyal2017something}. While these methods enhance temporal capacity from various perspectives, STSep consistently outperforms them across all configurations. Performance improves with more test clips (from 3 to 10), yielding more stable predictions and higher \texttt{Top-1/5} accuracy. Likewise, longer input sequences (8 to 16 frames) provide richer temporal context, further boosting discriminative power. These results confirm STSep's effective exploitation of temporal information.

Beyond superior results on the temporally demanding Something-Something V2, STSep generalizes well to UCF101\cite{soomro2012ucf101} and HMDB51\cite{kuehne2011hmdb}, as shown in Tables~\ref{tab:ucf101} and~\ref{tab:hmdb51}. This demonstrates the effectiveness of spatiotemporal separation across diverse scene types and limited-sample regimes. while minor gains for appearance-based bias, consistent improvements across these datasets, which vary in temporal complexity, scale, and action categories, validate STSep as a robust general temporal modeling strategy.

ImageNet pretraining delivers further gains over training from scratch with strong spatial feature prior information. This shows that STSep successfully transfers pretrained spatial knowledge to video tasks while simultaneously learning motion patterns through its temporal difference branch.

Remarkably, STSep with local dense sampling outperforms TSN's global sparse sampling, due to explicit temporal difference modeling that captures fine-grained motion. Moreover, combining STSep with TSN sampling yields more performance in short clips, offering a more comprehensive temporal representation for video understanding.

\subsection{Retrieval Task}
Retrieval tasks measure spatiotemporal feature extraction by computing similarity in high-dimensional space. Following procedure in \cite{han2019video}, we use validation sets from Something-Something V2\cite{goyal2017something}, UCF101\cite{soomro2012ucf101}, and HMDB51\cite{kuehne2011hmdb} as queries. We employ K-nearest neighbors (\textbf{KNN}) to evaluate \texttt{Recall@k} ($k\in\{1,3,5,10,20,50\}$), where retrieval succeeds if any \textit{top-k} neighbor matches the query class.

Table \ref{tab:retrieval} compares STSep and \texttt{vanilla} across three datasets with/without ImageNet pretraining. As expected, all models improve with larger $k$. STSep consistently outperforms \texttt{vanilla} across all configurations and \texttt{R@k} metrics, revealing stronger semantic feature extraction.

Remarkably, STSep without ImageNet pretraining surpasses \texttt{vanilla} with pretraining in SSV2, suggesting that spatialtemporal separation captures more discriminative motion features, while spatial prior knowledge may even hinder temporal modeling. This advantage is not pronounced on UCF101 and HMDB51, where weaker temporal dependencies make scene and object appearance more dominant.

Figure~\ref{fig:retrieval} visualizes \textit{top-3} retrievals for query samples. Despite different object appearances, the model correctly identifies similar actions like "pointing" or "diving" and clusters them semantically in feature space. This confirms that STSep successfully decouples object appearance from action semantics, focusing on temporal dynamics rather than static visual cues.

%% file: sec/5_conclusion.tex
\section{Conclusion}
\vspace{-0.5em}
We investigate SNNs' temporal modeling capability for video understanding. Through Non-Stateful ablation studies, we identify temporal modeling conflicts in SNNs and propose the Spatial Temporal Separable Network (STSep), which explicitly decouples temporal and spatial branch. STSep significantly improves temporal modeling while maintaining computational efficiency across multiple video datasets, offering an effective architectural paradigm for SNN-based temporal understanding tasks.

%% file: sec/X_suppl.tex
\clearpage
\setcounter{page}{1}
\setcounter{table}{7}
\setcounter{figure}{6}
\maketitlesupplementary

\section{NS ablation in UCF101/HMDB51}

In this section, we supplement Non-Stateful (\textbf{\texttt{NS}}) ablation experiments on UCF101 and HMDB51 datasets to verify the generality of the non-monotonic phenomenon observed on Something-Something V2. Specifically, maintaining identical settings as SSV2, we train a series of \texttt{NS} models on UCF101 and HMDB51, progressively removing temporal state at different stages and recording performance for each configuration. Results are shown in Table~\ref{tab:NS_appendix}.

We find that, consistent with observations in SSV2, moderate removal of temporal state in shallow or deep layers improves performance. However, further removal across more stages causes significant performance degradation. These results demonstrate that the observations are not random occurrences but rather consistent patterns observed across different types of datasets. Moreover, these results further support our spatiotemporal resource competition hypothesis and underscore the importance of adopting different temporal modeling strategies across network depths.

\begin{table}[h] 
\centering

\setlength{\tabcolsep}{11.5pt}
\renewcommand{\arraystretch}{1.0}
\scalebox{0.97}{ 

\begin{tabular}{l|cccc}
\textbf{datasets}       & \multicolumn{2}{c|}{\textbf{UCF101}}               & \multicolumn{2}{c}{\textbf{HMDB51}} \\ \hline
\textbf{input}          & \multicolumn{4}{c}{\textbf{$8\times128$}}                                                \\ \hline
\textbf{model}          & \textbf{Top1} & \multicolumn{1}{c|}{\textbf{Top5}} & \textbf{Top1}    & \textbf{Top5}    \\ \hline
\textbf{vanilla}        & 42.1          & \multicolumn{1}{c|}{68.6}          & 16.9             & 46.9             \\
\hdashline \textbf{NS1} & 43.0          & \multicolumn{1}{c|}{69.5}          & 17.1             & 47.1             \\
\textbf{NS2}            & 42.9          & \multicolumn{1}{c|}{70.5}          & 17.2             & 47.5             \\
\textbf{NS3}            & 42.4          & \multicolumn{1}{c|}{70.1}          & 16.5             & 47.2             \\
\textbf{NS4}            & 41.2          & \multicolumn{1}{c|}{67.6}          & 16.8             & 45.5             \\
\textbf{NS5/rNS5}       & 39.5          & \multicolumn{1}{c|}{66.7}          & 14.8             & 41.4             \\
\textbf{rNS4}           & 40.9          & \multicolumn{1}{c|}{66.8}          & 16.7             & 45.0             \\
\textbf{rNS3}           & 42.8          & \multicolumn{1}{c|}{70.2}          & 17.3             & 46.1             \\
\textbf{rNS2}           & 43.0          & \multicolumn{1}{c|}{69.6}          & \textbf{18.5}    & \textbf{47.9}    \\
\textbf{rNS1}           & \textbf{43.5} & \multicolumn{1}{c|}{\textbf{71.3}} & 18.1             & 47.1            
\end{tabular}

}
\caption{\textbf{NS ablation results on UCF101 and HMDB51 datasets.} Progressive removal of temporal state at different stages reveals non-monotonic performance trends, consistent with SSV2 observations.}
\label{tab:NS_appendix}
\vspace{-1.8em}
\end{table}

\section{Experimental Details}
In this section, we provide comprehensive details of experimental configurations, including \textbf{training hyperparameters}, \textbf{data preprocessing}, and \textbf{evaluation protocols}, to ensure our reproducibility of all results.

Across all experiments, we employ the AdamW optimizer with an initial learning rate of $6e^{-4}$, weight decay of $5e^{-6}$, and cosine annealing learning rate scheduling. The batch size is uniformly set to 256. Training proceeds for 50 epochs on Something-Something V2 and 100 epochs on UCF101 and HMDB51. All datasets adopt a spatial resolution of $128\times128$ pixels. All experiments are conducted on NVIDIA RTX 4090 GPUs with 4 GPUs per training session.

For data preprocessing, we apply random spatial cropping and horizontal flipping as augmentation strategies. Input frames are first resized such that the shorter side reaches $128\times1.2$ pixels, followed by random cropping to $128\times128$ pixels. Horizontal flipping is performed with probability 0.5, except for Something-Something V2 where temporal semantics preclude such augmentation. During evaluation, we adopt center cropping by resizing the shorter side to $128\times1.2$ pixels and extracting a central $128\times128$ region for inference.

\begin{table}[h] 
\centering

\vspace{-1.em}
\setlength{\tabcolsep}{7pt}
\renewcommand{\arraystretch}{1.0}
\scalebox{0.97}{ 

\begin{tabular}{lccc}
\multicolumn{1}{l|}{\textbf{dataset}}                                                        & \multicolumn{1}{c|}{\textbf{\begin{tabular}[c]{@{}c@{}}Something \\ SomethingV2\end{tabular}}} & \multicolumn{1}{c|}{\textbf{UCF101}} & \textbf{HMDB51} \\ \hline
\multicolumn{1}{l|}{\textbf{Epoch}}                                                          & \multicolumn{1}{c|}{50}                                                                        & \multicolumn{1}{c|}{100}             & 100             \\
\multicolumn{1}{l|}{\textbf{Batch Size}}                                                     & \multicolumn{1}{c|}{256}                                                                       & \multicolumn{1}{c|}{256}             & 256             \\
\multicolumn{1}{l|}{\textbf{LR}}                                                             & \multicolumn{1}{c|}{$6e^{-4}$}                                                                 & \multicolumn{1}{c|}{$6e^{-4}$}       & $6e^{-4}$       \\
\multicolumn{1}{l|}{\textbf{WD}}                                                             & \multicolumn{1}{c|}{$5e^{-6}$}                                                                 & \multicolumn{1}{c|}{$5e^{-6}$}       & $5e^{-6}$       \\
\multicolumn{1}{l|}{\textbf{RS}}                                                             & \multicolumn{1}{c|}{$128\times128$}                                                            & \multicolumn{1}{c|}{$128\times128$}  & $128\times128$  \\
\multicolumn{1}{l|}{\textbf{SyncBN}}                                                         & \multicolumn{1}{c|}{\checkmark}                                                                & \multicolumn{1}{c|}{\checkmark}      & \checkmark      \\ \hline
\multicolumn{4}{l}{\textbf{Data Augmentation}}                                                                                                                                                                                                         \\ \hline
\multicolumn{1}{l|}{\textbf{Resize}}                                                         & \multicolumn{1}{c|}{$128\times1.2$}                                                            & \multicolumn{1}{c|}{$128\times1.2$}  & $128\times1.2$  \\
\multicolumn{1}{l|}{\textbf{Crop}}                                                           & \multicolumn{1}{c|}{$128\times128$}                                                            & \multicolumn{1}{c|}{$128\times128$}  & $128\times128$  \\
\multicolumn{1}{l|}{\textbf{\begin{tabular}[c]{@{}l@{}}Horizontal \\ Flipping\end{tabular}}} & \multicolumn{1}{c|}{$p=0$}                                                                     & \multicolumn{1}{c|}{$p=0.5$}         & $p=0.5$        
\end{tabular}

}

\caption{\textbf{Detailed Experiments Details} on Something-Something V2, UCF101, and HMDB51 datasets. RS: Input Resolution; LR: Learning Rate; WD: Weight Decay; SyncBN: Synchronized Batch Normalization.}
\label{tab:experimental_details}
\vspace{-1.8em}
\end{table}

\section{More other Results}

\begin{table}[h] 
\centering

\vspace{-1.em}
\setlength{\tabcolsep}{1pt}
\renewcommand{\arraystretch}{1.0}
\scalebox{0.97}{ 

\begin{tabular}{l|l|c|cc}
\textbf{method}                                        & \textbf{model}                                             & \textbf{\begin{tabular}[c]{@{}c@{}}pretrain\\ weight acc\end{tabular}} & \multicolumn{1}{l}{\textbf{Top1}} & \multicolumn{1}{l}{\textbf{Top5}} \\ \hline
\textbf{Vanilla+TSN}                                   & SEW ResNet 18                                              & -                                                                      & 43.3                              & 70.8                              \\
\textbf{STSep}                                         & SEW ResNet 18                                              & -                                                                      & 48.5                              & 74.7                              \\
\textbf{STSep}                                         & SEW ResNet 18                                              & 63.2                                                                   & 69.5                              & 90.7                              \\
\hdashline \textbf{ReSpike} \cite{xiao2025respike}     & \begin{tabular}[c]{@{}l@{}}Hybird\\ (SNN+ANN)\end{tabular} & 73.2                                                                   & 77.5                              & 93.9                              \\
\textbf{SVFormer-st} \cite{yu2024svformer}             & SVFormer-st                                                & 82.9                                                                   & 80.2                              & -                                 \\
\textbf{STS ResNet} \cite{samadzadeh2023convolutional} & STS ResNet                                                 & -                                                                      & 42.1                              & -                                
\end{tabular}

}
\caption{\textbf{Comparison with other methods on UCF101.} Model performance varies significantly with different pretrained weights, therefore, we also report the accuracy of pretrained weights used by each method for reference.}
\label{tab:other}
\vspace{-1.8em}
\end{table}

\begin{figure*}[t]
    \centering 
    \textbf{Retrieval Results on UCF101}
    \includegraphics[width=1.0\linewidth]{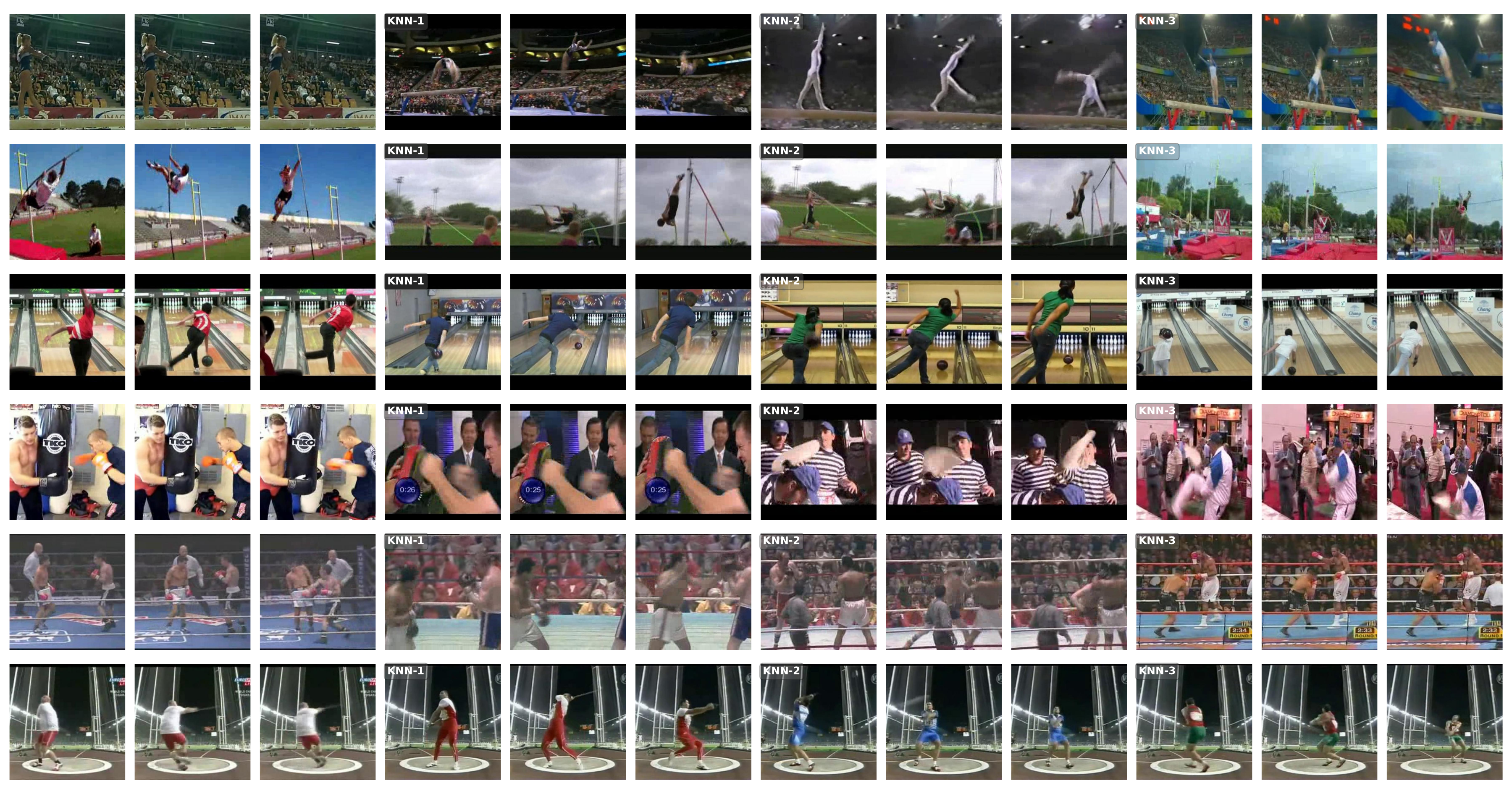}
    \caption{\textbf{More examples visualizations of video retrieval in UCF101.} Each query video sampled from UCF101 is shown with its \texttt{top-3} nearest neighbors in feature space. We extract features from the global average pooling layer before classifier, averaging across the temporal dimension for retrieval. }
    \label{fig:retrieval_ucf_appendix}
    \vspace{-1.em}
\end{figure*}

\begin{figure*}[t]
    \centering 
    \textbf{Retrieval Results on HMDB51}
    \includegraphics[width=1.0\linewidth]{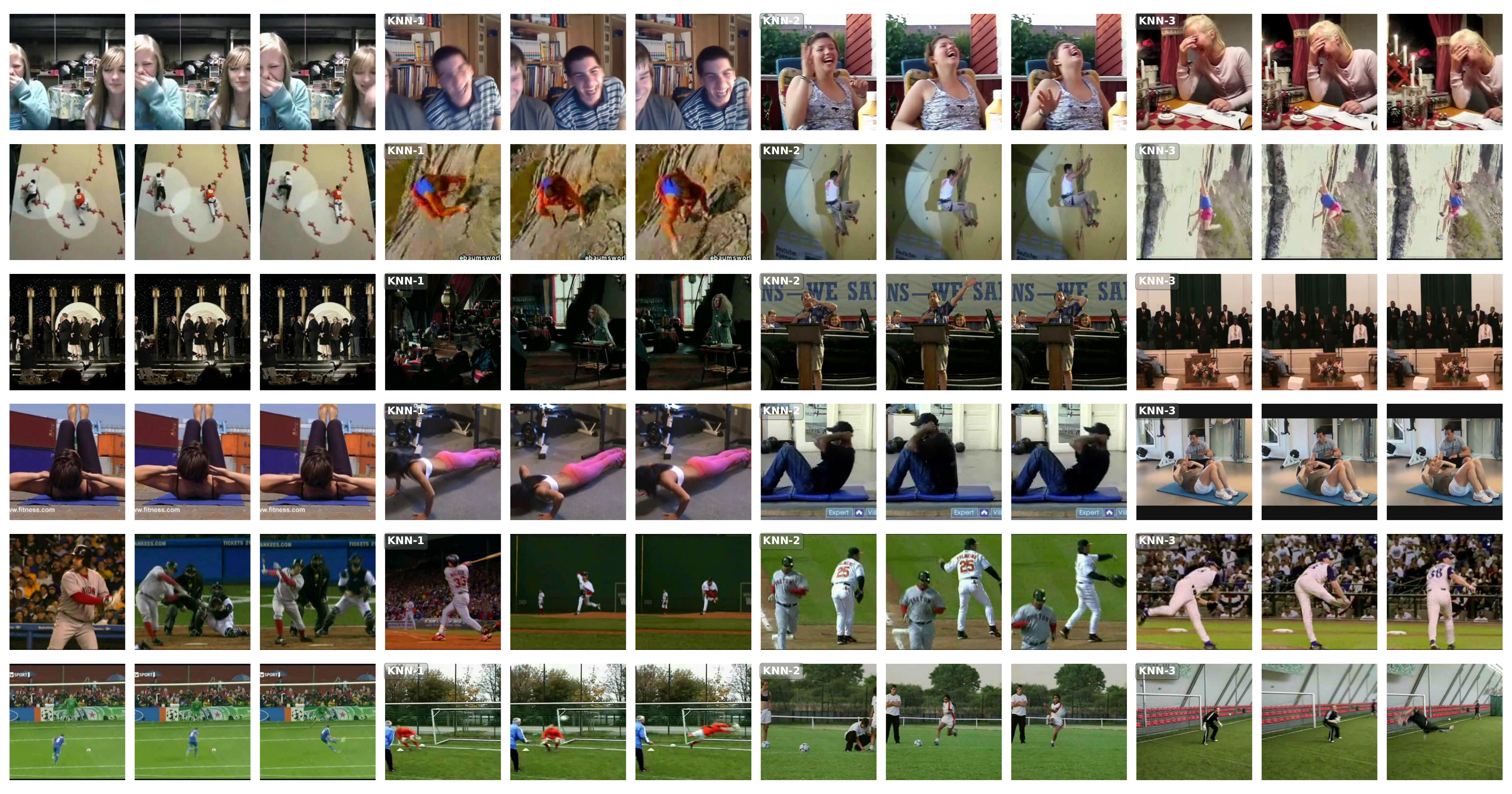}
    \caption{\textbf{More examples visualizations of video retrieval in HMDB51.} Each query video sampled from HMDB51 is shown with its \texttt{top-3} nearest neighbors in feature space. We extract features from the global average pooling layer before classifier, averaging across the temporal dimension for retrieval. }
    \label{fig:retrieval_hmdb_appendix}
    \vspace{-1.em}
\end{figure*}

\begin{figure*}[t]
    \centering 
    \includegraphics[width=0.99\linewidth]{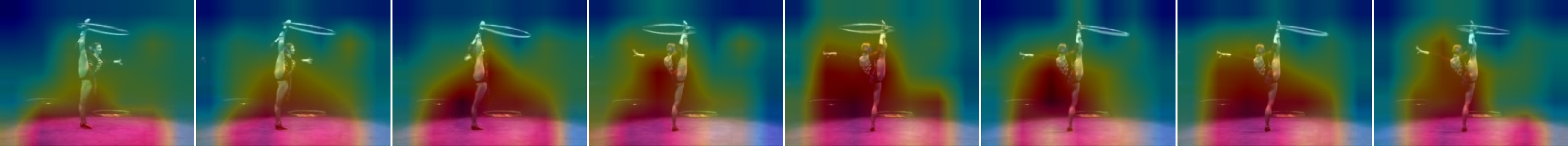}
    \includegraphics[width=0.99\linewidth]{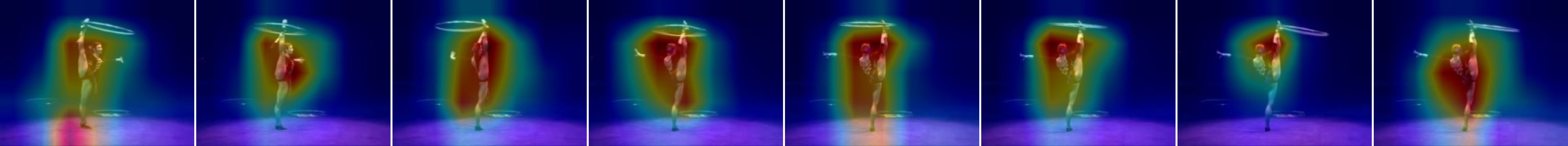}\\
    \vspace{1.em}
    \includegraphics[width=0.99\linewidth]{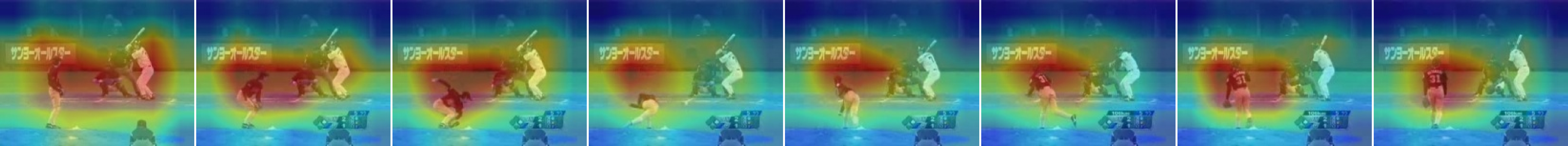}
    \includegraphics[width=0.99\linewidth]{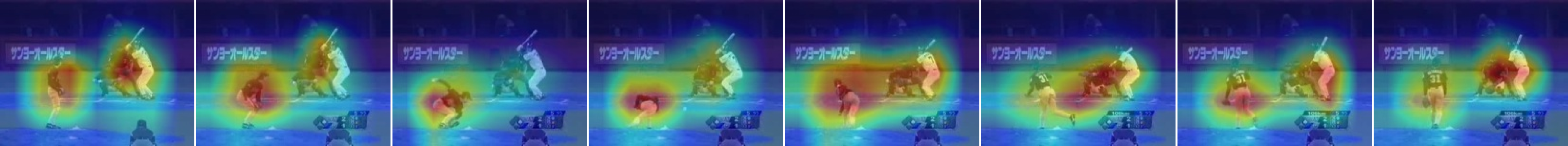}\\
    \vspace{1.em}
    \includegraphics[width=0.99\linewidth]{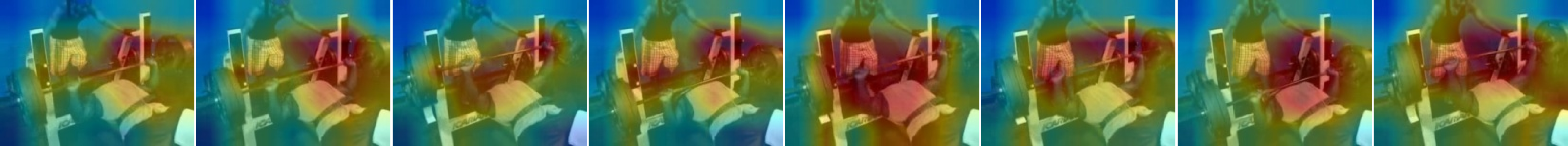}
    \includegraphics[width=0.99\linewidth]{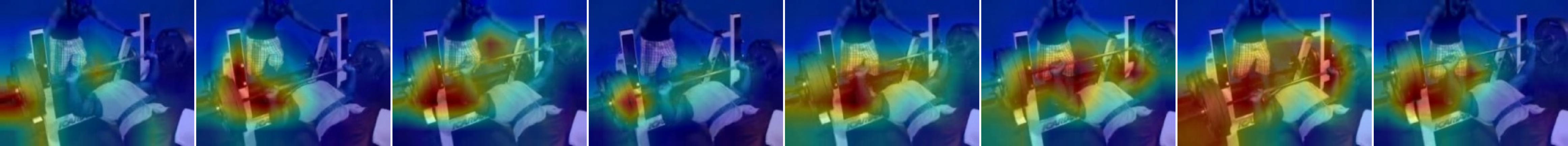}\\
    \vspace{1.em}
    \includegraphics[width=0.99\linewidth]{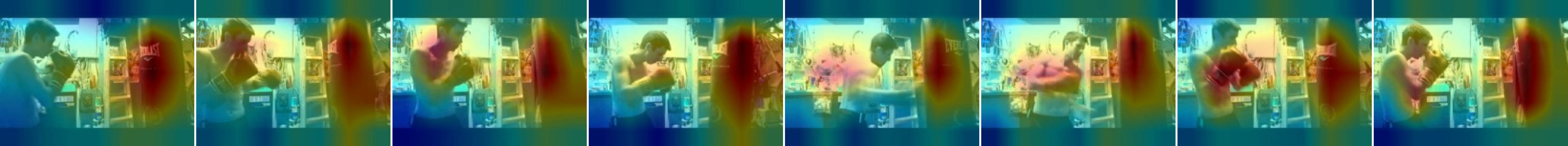}
    \includegraphics[width=0.99\linewidth]{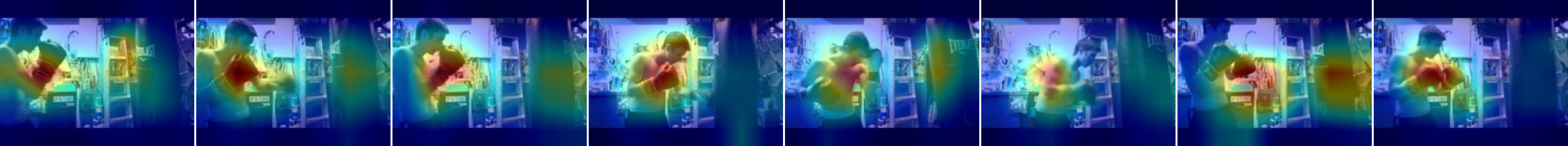}\\
    \vspace{1.em}
    \includegraphics[width=0.99\linewidth]{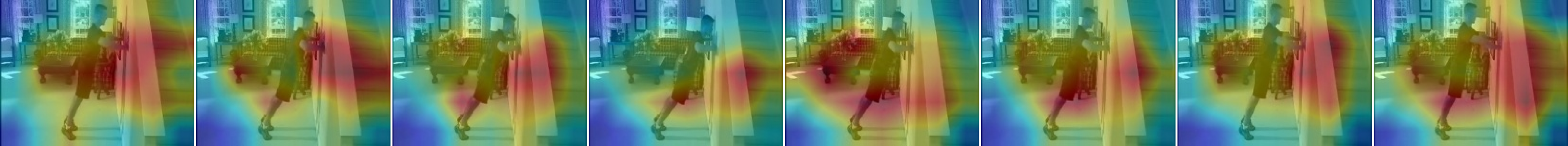}
    \includegraphics[width=0.99\linewidth]{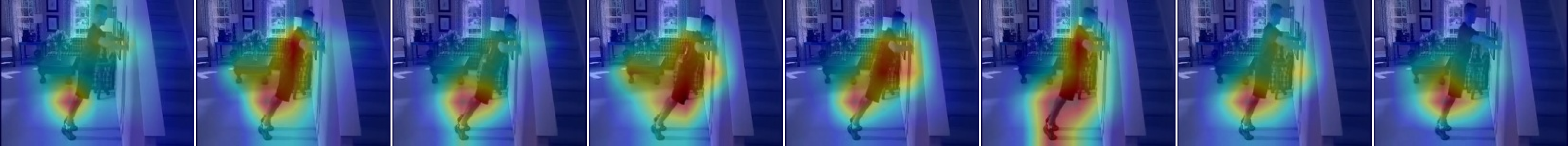}\\
    \vspace{1.em}
    \includegraphics[width=0.99\linewidth]{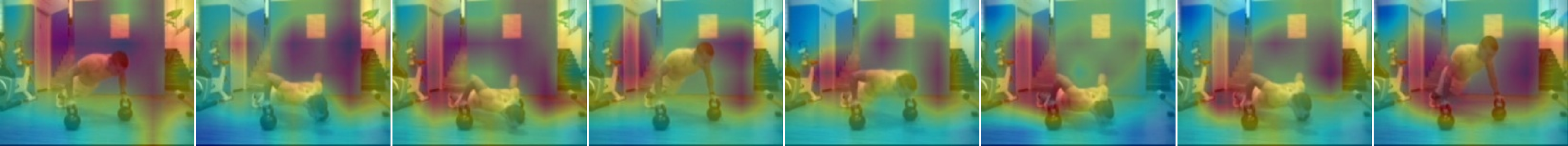}
    \includegraphics[width=0.99\linewidth]{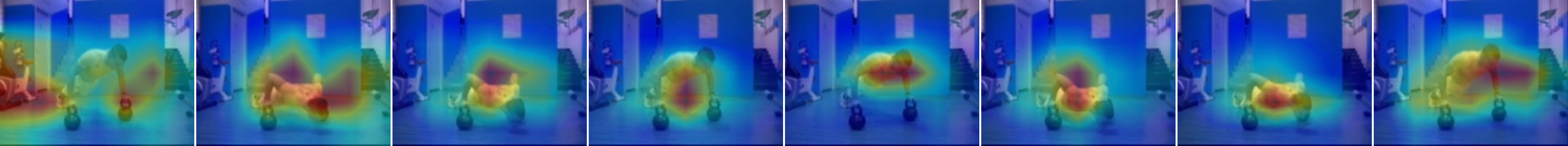}

    \caption{\textbf{More attention activations of video in UCF101.} Top: vanilla SNN; Bottom: STSep. STSep consistently focuses on motion regions rather than static appearance. }
    \label{fig:attn_appendix}
    \vspace{-1.em}
\end{figure*}

In this section, we compare additional methods evaluated on UCF101, including hybrid architectures, large-scale models with superior initialization weights, and others. Although direct comparison may not be feasible due to different experimental settings, we present these results to demonstrate STSep's effectiveness. On the other hand, we also include the accuracy of pretrained weights used by each method for reference, as model performance can vary significantly with different pretrained weights. The results are shown in Table~\ref{tab:other}.

\section{More Visualization Results}
In this section, Figure~\ref{fig:retrieval_ucf_appendix} and \ref{fig:retrieval_hmdb_appendix} presents additional visualization results on retrieval tasks to provide clearer observation of model performance.

\section{More Attention Map Visualization}
In this section, we provide extensive attention heatmap visualizations to comprehensively demonstrate STSep's effectiveness in attending to motion-relevant regions. We generate attention heatmaps using the method from \cite{kim2021visual}, which is a more effective model attention approach tailored for SNNs. We present additional comparative results between STSep and vanilla SNN on UCF101 and HMDB51 datasets. Figure~\ref{fig:attn_appendix} showcases more attention heatmap, further validating STSep's superiority in capturing dynamic information.